%% file: neurips_2026.tex
\documentclass{article}

\usepackage[numbers]{natbib}

\usepackage[main, final]{neurips_2026}

\usepackage[utf8]{inputenc} 
\usepackage[T1]{fontenc}    
\usepackage{url}            
\usepackage{booktabs}       
\usepackage{nicefrac}       
\usepackage{microtype}      
\usepackage[dvipsnames]{xcolor}         

\usepackage{amsthm, amssymb}
\usepackage{graphicx}
\usepackage{wrapfig}
\usepackage{subcaption}
\usepackage{algorithm}
\usepackage{algpseudocode}
\usepackage{booktabs}
\usepackage{multirow}
\usepackage{placeins}
\input{contents/math_commands.tex}

\usepackage[table]{xcolor}
\definecolor{paperhi}{RGB}{230,159,0}
\definecolor{citecolor}{RGB}{0, 113, 188}
\usepackage[pagebackref,breaklinks,colorlinks,citecolor=citecolor,linkcolor=red]{hyperref}

\title{SNACK: A Sequential Notation Framework for Probabilistic Graph Generation}

\author{%
Hohyun Kim\thanks{Equal contribution} \quad
Hyesung Kim\footnotemark[1] \quad
Min-hwan Oh\thanks{Corresponding authors. Emails: \texttt{\{minoh, lee7801\}@snu.ac.kr}} \quad
Seunggeun Lee\footnotemark[2]
\\ [0.3cm]
Seoul National University
}

\begin{document}

\maketitle
\setcounter{footnote}{0}

\begin{abstract}
We introduce SNACK, a string representation for probabilistic graph generation that enables language models to generate graphs with discrete node and edge features. SNACK linearizes graphs by encoding the nonzero entries of the lower-triangular adjacency matrix, establishing a bijective correspondence between valid sequences and ordered adjacency matrices. This design enables invalid logit masking to enforce structural constraints, such as chemical valence and ordering rules, thereby guaranteeing 100\% valid generated sequences while sticking to standard language modeling for effective training. SNACK also enables principled string-based GFlowNet training by explicitly accounting for node-ordering symmetries. Across standard graph-generation tasks and molecular distribution-learning benchmarks, SNACK achieves competitive sample quality, while substantially improving the throughput of GFlowNet training relative to graph-based generation. Probing analyses further show that models trained on SNACK encode meaningful local and global graph structure despite receiving only sequential supervision. These results position SNACK as a practical framework for probabilistic graph generation with deep sequence models.\footnote{Package available at \url{https://github.com/leelabsg/snack}}
\end{abstract}

\input{contents/body.tex}

\section*{Acknowledgments}
This research was supported by the Advanced GPU Utilization Support Program funded by the Korean government (Ministry of Science and ICT) (No. 02-26-01-0303), the Institute of Information \& Communications Technology Planning \& Evaluation (IITP) grant funded by the Korean government (MSIT) (No. RS-2025-25463302, Global Talent Recruitment for AI-powered Drug Discovery), the Korea Health Technology R\&D Project through the Korea Health Industry Development Institute (KHIDI) funded by the Ministry of Health \& Welfare, Republic of Korea (No. RS-2025-02307233), the National Research Foundation of Korea (NRF) grant funded by the Korean government (MSIT) (No. RS-2022-NR071853, RS-2023-00222663, RS-2026-25613008), the SNU AI-Bio Research Grant, and the Advanced Medical and Pharmaceutical Researcher Training Support Program through Seoul National University.

\bibliographystyle{unsrtnat}
\bibliography{contents/ref}
\newpage
\appendix
\input{contents/appendix.tex}



\end{document}

%% file: contents/math_commands.tex
\usepackage{amsmath,amsfonts,bm}

\newtheorem{proposition}{Proposition}
\newtheorem*{definition*}{Definition}

\newcommand{\Aut}{{\mathrm{Aut}}}

\definecolor{myblue}{RGB}{21, 118, 209}
\definecolor{mygreen}{RGB}{60,179,113}
\definecolor{mygrey}{RGB}{100, 100, 100}
\definecolor{fore}{RGB}{30, 30, 30}
\definecolor{back}{RGB}{255,255,255}
\definecolor{orange}{rgb}{0.85,0.33,0.13} 
\definecolor{lime}{rgb}{0.65,0.85,0.13}
\definecolor{bred}{RGB}{203,65,84}

\def\green#1{\textcolor{mygreen}{#1}}

\def\eqref#1{equation~\ref{#1}}

\def\1{\bm{1}}

\DeclareMathAlphabet{\mathsfit}{\encodingdefault}{\sfdefault}{m}{sl}
\SetMathAlphabet{\mathsfit}{bold}{\encodingdefault}{\sfdefault}{bx}{n}



%% file: contents/body.tex
\section{Introduction}

Two dominant paradigms have been widely explored for generative models for graphs with discrete features: graph-based methods \citep{li2018learning,you2018graph,you2018graphrnn}, which operate directly on discrete structures, and string-based methods \citep{krenn2020self,ahn2021spanning,jang2023simple}, which linearize graphs into sequences. While graph-based methods naturally handle valid actions or trajectories, they often demand specialized architectures and incur high computational costs. In contrast, string-based representations enable efficient training using standard sequence models. However, existing string-based methods fall short as flexible graph learning frameworks, as they typically address only a subset of key capabilities, such as guaranteed validity, general graph applicability, flexible conditional generation, or goal-directed exploration. Furthermore, a single graph can correspond to multiple valid string representations, which poses a significant challenge to probabilistic modeling. While this issue has been extensively studied in graph-based frameworks \citep{chen2021order, kim2025symmetry}, it has received comparatively little attention in the context of string-based approaches. Consequently, graph-based methods remain the predominant choice in frameworks such as Generative Flow Networks (GFlowNets) \citep{bengio2021flow}.

In this work, we address these challenges by introducing \textbf{S}equential \textbf{N}otation of \textbf{A}djacent \textbf{C}onnectivity Framewor\textbf{K} (SNACK), a novel string representation that addresses the aforementioned challenges by enabling flexible graph learning. The encoding process of SNACK mirrors the iterative construction found in adjacency matrix-based methods, where a graph is linearized row-wise. This design allows SNACK to inherit core advantages of graph-based methods, while utilizing efficient sequence training pipelines. Specifically, our learning framework provides three practical advantages: first, each representation decodes into a unique adjacency matrix, facilitating string-based GFlowNet training by introducing a node ordering policy; second, any sequence prefix can be readily decoded into a partial graph state, which enables conditioned optimization and direct logit masking during autoregressive generation; and third, SNACK maintains a compact representation, utilizing exactly $N + 2M$ tokens for an attributed graph with $N$ nodes and $M$ edges, thereby avoiding redundancy while preserving structural correspondence. 

We demonstrate the efficacy of SNACK across various graph-related tasks, including distribution learning and property optimization. Our results show that SNACK reliably generates high-quality samples, achieving 100\% validity across all benchmarks. Furthermore, SNACK facilitates the implementation of string-based GFlowNets, significantly enhancing the scalability of GFlowNet training on graphs. We also introduce a conditioned Trajectory Balance (TB) loss specifically tailored for constrained generation. Beyond performance gains, we provide empirical evidence that SNACK internalizes graph structural information within its representation despite its sequential nature, a finding that is of independent interest. We summarize our contributions as follows.

\begin{itemize}
    \item We introduce SNACK, a simple and efficient string representation that maintains a one-to-one correspondence with adjacency matrices. 
    \item We propose a training methodology for string-based GFlowNets applicable to both \textit{de novo} and conditioned settings, which significantly improves the scalability.
    \item We evaluate SNACK both on distribution and goal-directed learning tasks, showing competitive generation quality, guaranteed 100\% validity, and superior exploration capabilities.
    \item We employ bilinear probing to show that SNACK-based models internalize graph structures within their representations from purely sequential data without explicit supervision.
\end{itemize}

\section{Sequential Notation of Adjacent Connectivity Framework}

This section introduces the SNACK framework. We analyze its core properties and define the training objectives necessary for distribution learning and GFlowNet.

\subsection{Overview}
\label{sec:overview}

SNACK follows a similar procedure to adjacency matrix-based generative models for undirected graphs, where a graph is constructed node-by-node, along with its corresponding edges to existing nodes. For instance, \texttt{2=1-[C]} specifies that a node of type \texttt{[C]} is connected to the nodes located two positions and one position earlier, using edge types \texttt{=} and \texttt{-}, respectively. 


Formally, consider an undirected graph \(G=(V, E)\) in which each node and edge is associated with a type label. To encode the graph, we assign a unique index to each node via a mapping \(\pi : V \rightarrow [1, N]\). Let \(v_i\) denote the label of the \(i\)-th node, and let \(e_{i,j}\) denote the label of the edge between nodes \(i\) and \(j\). We define the distance between two nodes \(i\) and \(j\) as \(|i-j|\). If the \(k\)-th node is connected to nodes \(i\) and \(j\) with \(i < j < k\), the corresponding encoding is
\[
[\; \dots,\; 
\textcolor{BrickRed}{\bm{v_i}},\quad  \dots,\; 
\textcolor{BrickRed}{\bm{v_j}},\quad  \dots,\quad 
\textcolor{ForestGreen}{\bm{|k-i|}},\quad  
\textcolor{YellowOrange}{\bm{e_{k,i}}},\quad 
\textcolor{ForestGreen}{\bm{|k-j|}},\quad  
\textcolor{YellowOrange}{\bm{e_{k,j}}},\quad 
\textcolor{BrickRed}{\bm{v_k}} 
\; ]
\]
where \(|k-i|\) and \(|k-j|\) represent the distances from node \(k\).

\begin{figure} 
    \centering
    \includegraphics[width=\linewidth]{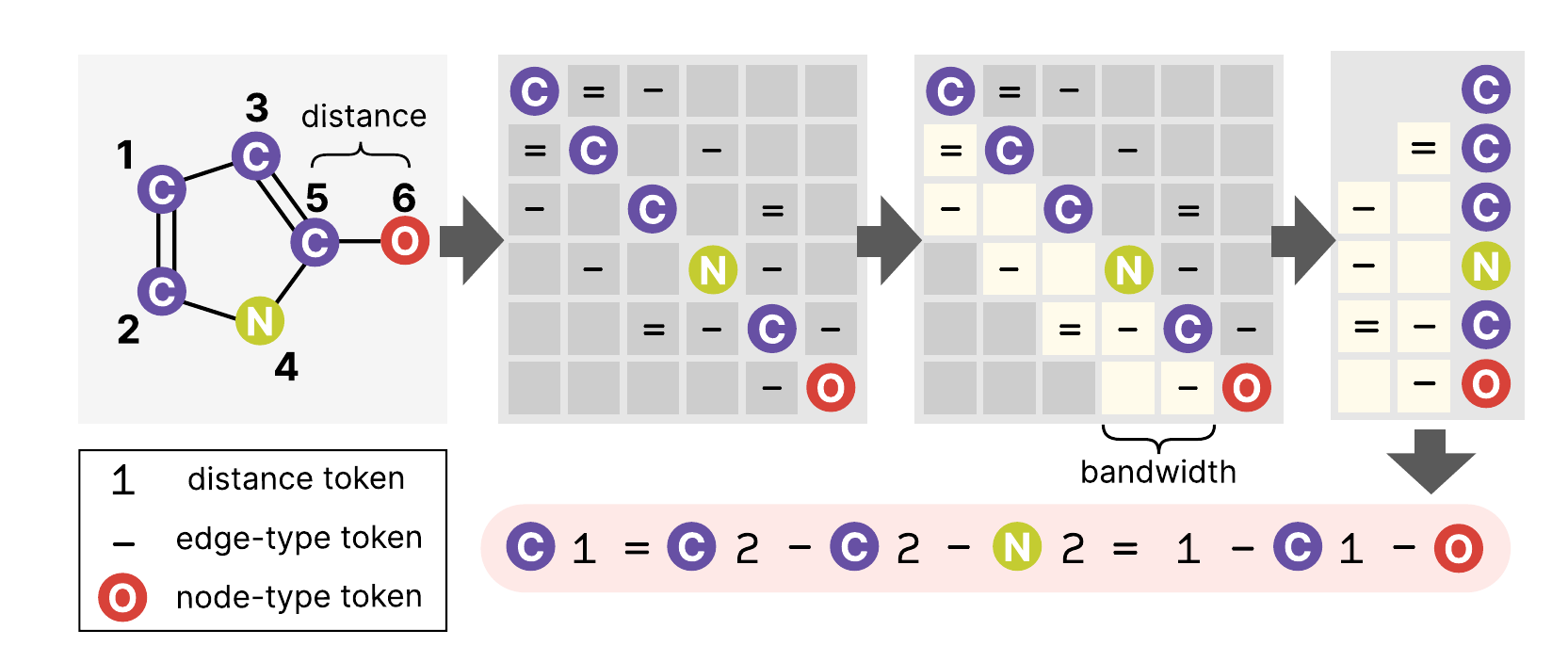}
    \caption{Overview of the encoding process. The molecule in the figure is represented as \texttt{[C]1=[C]2-[C]2-[N]2=1-[C]1-[O]}. By reordering the node indices, the same graph can also be represented as \texttt{[O][C][N]3-2=1-[C]3-[C]3-1=[C]}, resulting in a larger bandwidth.}
    \label{fig:main diagram}
    \vspace{-10pt} 
\end{figure}

We derive the encoding procedure from the graph's adjacency matrix, as illustrated in Figure~\ref{fig:main diagram}. The adjacency matrix $\mathbf{A}$ is defined as a symmetric $N \times N$ matrix, where off-diagonal entries specify edge types and diagonal entries specify node types. Since all structural information is contained in the lower triangular part of the matrix, it suffices to encode only the lower triangular portion, including the diagonal. We encode the graph row by row in ascending order of node IDs, recording non-zero entries. We adopt the following rules to encode each row $k$: (1) each off-diagonal entry $(k, i)$ is encoded as two tokens, $(k-i)$ and $e_{k,i}$, while the node itself is represented by its diagonal entry $v_k$; (2) within each row, non-zero entries are encoded from left to right. Left-to-right order prioritizes neighbors that were introduced earlier, thereby reducing the required context window during decoding. We evaluate other design choices for SNACK in the experimental section.

\subsection{Bandwidth-restricted encoding}

In the current scheme, the distance between two nodes can be as large as \(N-1\), which requires \(N-1\) distinct distance tokens in the vocabulary. To reduce the vocabulary size, we aim to keep distances small by choosing an appropriate node ordering \(\pi\). The largest distance induced by an ordering is known as its \emph{bandwidth}. As illustrated in Figure~\ref{fig:main diagram}, the ordering function \(\pi\) can be chosen so that all nonzero entries are concentrated within a narrow band around the diagonal.

Restricting the bandwidth of the adjacency matrix has been shown to substantially improve the performance of graph generative models \citep{diamant2023improving,jang2023simple}. Random breadth-first search (BFS) ordering is a standard approach for bandwidth reduction \citep{you2018graphrnn,popova2019molecularrnn,shi2020graphaf,luo2021graphdf}, while the Cuthill--McKee (CM) algorithm \citep{cuthill1969reducing}, a BFS variant, has been proposed to further minimize bandwidth \citep{diamant2023improving}. However, in our molecular experiments, BFS ordering consistently outperforms CM. We therefore adopt BFS ordering to control the vocabulary size in SNACK. See Appendix~\ref{app:different node orderings} for a detailed discussion.

\subsection{Invalid logit masking}
\label{sec:invalid logit masking}
Each element in a SNACK sequence corresponds directly to a component of the graph, such as a node or an edge, making the representation easy to interpret. As a result, an intermediate graph can be readily decoded from a partially generated sequence. This prefix-decodability is central to controlled generation: at each step, we are allowed to inspect the partial graph to determine the exact set of admissible next operations, such as valid node connections or compliance with BFS ordering rules. We then convert these operational constraints into a logit mask applied directly to the model's output distribution, strictly preventing invalid sequence completions before token selection. Specifically, we employ three masking schemes:
\begin{itemize}
\item \textbf{Grammar:} We mask tokens that violate syntactic constraints, ensuring that the generated graphs are always valid SNACK representations. This includes enforcing termination by allowing only the \texttt{[eos]} token when appropriate, preventing infinitely long sequences.
\item \textbf{Valence:} For molecular graphs, we mask tokens that violate chemical validity, ensuring that the generated molecules are always chemically valid.
\item \textbf{BFS:} We mask tokens that do not respect the BFS ordering. 
\end{itemize}
We provide the rules for enforcing BFS masking in Appendix~\ref{app:Valid masking rule}. The BFS mask is crucial, as restricting the training examples to a specific ordering does not guarantee that the model will adhere to the same ordering during generation. Enforcing BFS ordering reduces the modeling space and maintains consistency with the training data, which helps mitigate exposure bias \citep{ranzato2015sequence}. To avoid confusion, note that BFS masking restricts the set of valid vocabulary tokens the model can predict, whereas BFS ordering determines the training sequence in which the graph is linearized.

\subsection{Sequence modeling objective}
\label{sec:training objective}
Let a tokenized sequence be denoted as $\mathbf{x} = [x_1, x_2, \dots, x_L]$, where $x_i$ is the $i$-th token. The negative log-likelihood of a sequence under model $p_\theta$ is given by
\begin{equation}
    \label{eqn:negative log-likelihood}
    \mathcal{L}_{\mathrm{NLL}}(\mathbf{x}; \theta) = -\log p_\theta(\mathbf{x}) = -\sum_{i=1}^{L} \log p_\theta(x_i|\mathbf{x}_{<i}),
\end{equation}
where the next-token probability is conditioned on the previous tokens $\mathbf{x}_{<i}$, and invalid tokens are masked out. 
The distribution under which the loss is computed is induced by the graph data and the node ordering policy. Let $q(\mathbf{x}|G)$ denote a distribution over sequences given a graph $G$. As noted in several prior works \citep{li2018learning,chen2021order,chen2025graph}, Equation~\ref{eqn:negative log-likelihood} optimizes a variational lower bound on the graph log-likelihood, given by
\begin{equation}
    \label{eqn:graph log-likelihood}
    \log p_\theta(G) \ge \mathbb{E}_{\mathbf{x} \sim q(\mathbf{x}| G)} \left[ \log p_\theta(\mathbf{x}) - \log q(\mathbf{x}| G) \right].
\end{equation}
While we focus on the setting where $q$ is fixed to a random BFS ordering throughout this paper, the following section describes how to explicitly compute $q$, as it is essential for GFlowNet training.

\subsection{Computing the variational distribution}

To compute $q(\mathbf{x}|G)$, we first note the one-to-one mapping between adjacency matrices and SNACK representations. That is, each SNACK sequence uniquely determines an ordered adjacency matrix, and each such matrix has a unique SNACK representation (proof in Appendix~\ref{app:bijectivity}).

\begin{proposition}
    \label{prop:bijectivity}
    The SNACK representation is bijective with respect to the adjacency matrix.
\end{proposition}

This implies that $q(\mathbf{x}|G)$ can be computed by evaluating the probability assigned to the corresponding adjacency matrix $\mathbf{A}$. This property distinguishes SNACK from other string-based representations, in which multiple sequences may map to the same adjacency matrix. For instance, in SELFIES, the strings \texttt{[C][N]} and \texttt{[Ring1][C][N]} yield the same adjacency matrix. 

Let $q^{\mathrm{ord}}(\pi|G)$ denote the ordering policy (e.g., random BFS ordering) that assigns probability to each node ordering $\pi$. While a node ordering determines an adjacency matrix $\mathbf{A}$ (and hence a SNACK sequence $\mathbf{x}$), multiple node orderings may induce the same matrix $\mathbf{A}$. Consequently, $q(\mathbf{x}|G)$ is obtained by summing $q^{\mathrm{ord}}(\pi|G)$ over all node orderings that preserve $\mathbf{A}$. The set of all node orderings that result in the same adjacency matrix is known as the automorphism group of the matrix, denoted by $\Aut(\mathbf{A})$. Thus, $q(\mathbf{x}|G)$ is given by:
\vspace{-5pt}
\begin{equation}
    q(\mathbf{x}|G) = \sum_{\pi\in\Aut(\mathbf{A})} q^{\mathrm{ord}}(\pi|G) = |\Aut(\mathbf{A})| \cdot q^{\mathrm{ord}}(\pi|G).
\end{equation}
The second equality holds because the ordering policy assigns equal probability to all node orderings that yield $\mathbf{A}$. This is a natural consequence of symmetry: node orderings within the same automorphism group induce the same adjacency matrix and are thus structurally indistinguishable.

The ability to evaluate $q$ opens up several possibilities, including: (1) training GFlowNets, (2) substructure-conditioned optimization, (3) learning $q$, and (4) computing the marginal log-likelihood. We focus specifically on GFlowNets in the next section and discuss the remaining applications in detail in Appendix~\ref{app:graph-to-seq_transition_probability}.

\subsection{Training GFlowNets with SNACK}
\label{sec:training gflownets}

Beyond generating graphs that resemble the training examples, property optimization aims to generate graphs with desired properties. To this end, GFlowNets have gained popularity, as they encourage the generative model to sample high-reward molecular graphs while maintaining diversity \citep{bengio2021flow}. Previous GFlowNet approaches have typically relied on graph-based representations \citep{bengio2021flow,malkin2022trajectory,jain2023multi,pandey2025pretraining,kim2025symmetry}, possibly because computing $q(\mathbf{x}|G)$ for string-based representations is challenging. Although SMILES-based GFlowNets have been adopted owing to the simplicity of string formats \citep{kim2024genetic,kim2026synthesizable,wang2026rooted}, this formulation inherently biases generation toward molecules with multiple valid SMILES strings \citep{bengio2021flow}. Consequently, how to properly align string representations with GFlowNets remains largely unexplored. In this section, we demonstrate that GFlowNets can be successfully trained using SNACK.

Let $\mathcal{X}$ denote the set of all SNACK representations and $\mathcal{X}_G$ denote the set of all SNACK representations of graph $G$. The marginal probability of generating $G$ is then given by $\sum_{\mathbf{x} \in \mathcal{X}_G} p_\theta(\mathbf{x})$. The goal of GFlowNets is to learn a policy such that $\sum_{\mathbf{x} \in \mathcal{X}_G} p_\theta(\mathbf{x}) = R(G) / Z$, where $R: \mathcal{G} \to \mathbb{R}_{>0}$ denotes a positive reward defined over the set of graphs $\mathcal{G}$, $G$ is the graph decoded from any sequence $\mathbf{x} \in \mathcal{X}_G$, and $Z$ is a normalizing constant. 

We define the state space as $\mathcal{S} = \mathcal{X} \cup \mathcal{G}$. Starting from the empty string (or the \texttt{[bos]} token), the policy generates the next token $x_i$ with probability $p_\theta(x_i|\mathbf{x}_{<i})$. Once the sequence is complete, a reward $R(G)$ is given, where $G$ is the graph decoded from the sequence $\mathbf{x}$. A complete forward trajectory is thus given by $[\mathbf{x}_{\leq 1}, \dots, \mathbf{x}_{\leq L-1}, \mathbf{x}, G]$, where the transition from $\mathbf{x}$ to $G$ is deterministic. Training via GFlowNets also requires evaluating backward trajectory probabilities. In our setting, intermediate backward transitions $\mathbf{x}_{\leq i+1} \to \mathbf{x}_{\leq i}$ are deterministically defined by removing the last token, while the initial step $G \to \mathbf{x}$ is governed by $q(\mathbf{x}|G)$. Under this formulation, the Trajectory Balance (TB) objective \citep{malkin2022trajectory} matches two trajectory distributions, $p_\theta(\tau)$ and $q(\mathbf{x}|G)R(G)/Z$, by minimizing squared error in the log domain:
\begin{eqnarray}
    \label{eqn:gflownet_objective}
    \mathcal{L}_{\mathrm{TB}}(\mathbf{x}, G;\theta, \gamma)
    =
    \left[
    \log Z_\gamma
    + \log p_\theta(\mathbf{x})
    - \log q(\mathbf{x}|G)
    - \log R(G)
    \right]^2,
\end{eqnarray}
where $\gamma$ is jointly optimized with $\theta$. The role of $q(\mathbf{x}|G)$ is to distribute the reward across all SNACK representations corresponding to $G$. The TB loss aims to train the model that samples trajectories in proportion to their assigned rewards, such that $p_\theta(\mathbf{x}) \propto q(\mathbf{x}|G)R(G)$, where $\log Z_\gamma$ acts as a variance reduction term \citep{malkin2022gflownets}. We further develop a conditioned version of GFlowNets that integrates initial state distributions for predefined substructures. The full derivation is provided in Appendix~\ref{app:conditioned gflownet}.

We use random BFS ordering, for which $q^{\mathrm{ord}}(\pi|G)$ can be evaluated during the BFS traversal, as detailed in Algorithm~\ref{alg:random_bfs}. Since our logit masking scheme ensures that the generation process adheres to a BFS ordering, the supports of $p_\theta$ and $q$ are perfectly aligned. This alignment guarantees that every generated sequence $\mathbf{x}$ possesses a positive backward flow, $q(\mathbf{x}|G)R(G)$. In contrast, models that can generate invalid strings (such as SMILES-based approaches) typically penalize invalid sequences by assigning an arbitrarily low value to $\log q(\mathbf{x}|G)$ to suppress their generation, which potentially leads to numerical instability.

\subsection{Overall computation}

Although computing $|\Aut(\mathbf{A})|$ is asymptotically expensive \citep{babai1983computational}, in practice, the graphs considered in existing generative models are typically limited to a few hundred nodes. This allows the \textit{nauty} package \citep{mckay2025nauty} to compute $|\Aut(\mathbf{A})|$ efficiently within a fraction of a millisecond (see Appendix J of \citet{kim2025symmetry}). This cost is also inherent to graph-based GFlowNets, which must evaluate $|\Aut(\mathbf{A})|$ to account for equivalent actions \citep{kim2025symmetry}.

For attributed graphs, SNACK uses $N + 2M$ tokens, and its vocabulary size scales linearly with bandwidth. SNACK computes trajectory-level log-probabilities in a single forward pass over the full sequence. In contrast, graph-based methods evaluate action probabilities at each intermediate state, requiring $N+M$ forward passes. This leads to higher computational cost and memory usage due to repeated processing of intermediate states. When both methods use standard dense attention \citep{vaswani2017attention}, this corresponds to a reduction in both computational and memory complexity from cubic to quadratic in the number of nodes (see Appendix~\ref{app:computational_complexity} for details).

\section{Related work}

\paragraph*{Graph generative models}
GraphRNN \citep{you2018graphrnn} generates graphs by constructing the adjacency matrix row-by-row. While GraphRNN explicitly predicts the presence or absence of every potential edge, our proposed method, SNACK, only encodes non-zero entries, significantly improving efficiency for sparse graphs. The utility of graph bandwidth in improving generative performance was first explored by \citet{diamant2023improving}. Our treatment of symmetry aligns with \citet{chen2021order}, who derived an identical automorphism formula for adjacency matrix-based graph generation. While invalid logit masking is commonly utilized in graph-based models to enforce chemical valency \citep{liu2018constrained, popova2019molecularrnn, shi2020graphaf}, these techniques primarily focus on improving the validity of generated molecules rather than reducing the output space. Similarly, recent string-based approaches like GEEL \citep{jang2023simple} and STGG \citep{ahn2021spanning} employ logit masking to satisfy syntactic and chemical constraints. Our work advances this line of research by enforcing a strict BFS ordering during generation through masking.

\paragraph*{String representations of graphs}
For molecular graphs, SMILES \citep{weininger1988smiles} is the most widely adopted string representation. Despite being human-readable, SMILES often suffers from low syntactic validity in generative tasks, such as unmatched ring closures. To address this, several robust alternatives have been proposed, such as DeepSMILES \citep{o2018deepsmiles}, SELFIES \citep{krenn2020self}, and STGG \citep{ahn2021spanning}. Recent methods such as G2PT \citep{chen2025graph}, GEEL \citep{jang2023simple} and AutoGraph \citep{chen2025flatten} have also been proposed as general edge list-based graph string representations. Specifically, GEEL uses CM ordering to restrict the vocabulary size; however, it differs from ours in that it constructs graphs edge by edge, and its vocabulary size scales quadratically with the bandwidth. While these approaches are similar to SNACK in their use of graph linearization, their scope and primary objectives fundamentally differ from our work. SNACK is specifically designed to facilitate effective goal-directed learning by integrating graph symmetry and specialized masking schemes. While certain existing methods could potentially be extended to achieve bijectivity relative to the adjacency matrix, to the best of our knowledge, we are the first to explicitly formalize and realize this bijective property within a string-based representation, providing a rigorous foundation for GFlowNet training. See Appendix~\ref{app:discussion} for further comparisons.


\paragraph*{GFlowNets}
While prior GFlowNet applications in molecular design have predominantly relied on graph-based approaches \citep{bengio2021flow, jain2022biological, jain2023multi, pandey2025pretraining}, we adopt string-based representations to substantially improve scalability. Consistent with recent findings that exact graph GFlowNets require automorphism correction \citep{kim2025symmetry}, our method integrates graph symmetries to guarantee unbiased sampling. Existing string-based GFlowNets operate on natural language \citep{hu2023amortizing,kwon2026stable,kim2026beyond} or SMILES \citep{kim2024genetic,kim2026synthesizable,wang2026rooted}; however, SMILES encoding inherently biases the model toward over-sampling larger molecules \citep{bengio2021flow}. Our formulation moves beyond standard autoregressive generation by inducing a directed acyclic graph (DAG) over the state space, allowing multiple distinct strings to transition to the same terminal graph.

\section{Experiments}
\label{sec:experiments}

In this section, we evaluate SNACK in three settings: generic graph generation, molecular string generation, and goal-directed learning, with a particular focus on molecular generation as a practically relevant application. For distribution learning, we employ 8.9M-parameter, 6-layer Transformers \citep{vaswani2017attention} based on the LLaMA architecture \citep{touvron2023llama}. For goal-directed learning tasks, a more compact 4-layer variant is utilized. Further implementation and experimental details are provided in Appendix~\ref{app:experimental details}.

\subsection{Generic graph generation}

We evaluate SNACK on eight widely used generic graph datasets that include both synthetic and real graphs, enabling us to assess its ability to model diverse graph distributions. Since these are non-attributed graphs, we omit edge-type tokens in the SNACK representation. Following \citet{you2018graphrnn}, we use squared Maximum Mean Discrepancy (MMD), which compares the distributions of graph statistics between generated and test graphs. We use the same set of hyperparameters across all datasets without dataset-specific tuning. Table~\ref{tab:generic_main} presents a subset of our results, with the full experimental data available in Appendix~\ref{app:additional_experimental_results}. SNACK generates samples whose statistics closely match the training distribution, approaching the benchmark's performance ceiling. These findings demonstrate SNACK's ability to model diverse graph structures without dataset-specific tuning.

\input{contents/table/generic_main}

\subsection{Molecular string generation}

\input{contents/table/mol_table}

In this section, we focus our comparison specifically on molecular strings using the QM9 \citep{ramakrishnan2014quantum} and MOSES \citep{polykovskiy2020molecular} dataset. We compare SNACK against established molecular string representations, such as SMILES \citep{weininger1988smiles}, SELFIES \citep{krenn2020self}, STGG \citep{ahn2021spanning}, and SAFE \citep{noutahi2024gotta}, alongside domain-agnostic methods including GEEL \citep{jang2023simple} and G2PT \citep{chen2025graph}. These baselines encompass both BFS- and DFS-based formulations. For our method, we evaluate three logit masking variants: no masking (None), grammar and valence masking (V), and a combination of grammar, valence and BFS masking (V+B). We report standard distribution-learning metrics, as proposed in the MOSES benchmark \citep{polykovskiy2020molecular}. By standardizing the underlying architecture and excluding method-specific tricks, we ensure that any observed performance gains are directly attributable to the representations themselves. Additional comparisons across more baselines and metrics are provided in Appendix~\ref{app:additional_results_moses}.

SNACK achieves competitive performance in molecular distribution learning across both datasets. While it remains strong even without masking, adding logit masking enables SNACK to reach 100\% validity. This effect is even more pronounced on the more challenging Peptides-func dataset \citep{dwivedi2022long}, as detailed in Appendix~\ref{app:peptides}. Overall, SNACK proves to be a highly competitive string-based method for graph distribution learning. Nevertheless, other string representations such as SMILES, SELFIES, and GEEL perform within one standard deviation across several metrics, achieving state-of-the-art performance. While differences among representations remain meaningful, all string-based methods benefit from the strong capabilities of language modeling.

\begin{wraptable}{r}{0.25\textwidth}
\vspace{-12pt}
\centering
\caption{Success Rate in conditioned generation.}
\label{tab:conditioned_succ}
\begin{tabular}{lc}
\toprule
Method & Success \\
\midrule
SMILES       & 0.152 \\
SELFIES      & 0.181 \\
SAFE         & 0.057 \\
STGG         & 0.003 \\
GEEL         & 0.000 \\
G2PT         & 0.000 \\
\midrule
SNACK & \cellcolor{paperhi!15}\textbf{0.390} \\
\bottomrule
\end{tabular}
\vspace{-10pt}
\end{wraptable}

The representational properties of SNACK are most clearly demonstrated in conditioned generation. The task is to continue the generation from given test fragments using a pretrained model without fine-tuning. For the 129 smallest MOSES test molecules, 100 random sequences were provided as inputs to the pretrained models. In most cases, baseline models tend to terminate immediately. In contrast, SNACK (None) yields a high success rate (Table~\ref{tab:conditioned_succ}), where success requires a generated molecule to be valid, connected, contain the target substructure, and possess strictly more heavy atoms. We attribute this to two factors: (1) SNACK allows the model to flexibly attach new atoms to any prior atom within the bandwidth, much like graph-based methods, whereas SMILES can only extend the sequence from the final atom. (2) For SNACK, every subsequence ending with a node-type token constitutes a valid prefix, thereby effectively yielding a larger training set for this task. We discuss this further in Appendix~\ref{app:substructure_conditioned_generation}.


\subsection{Goal-directed learning}

\input{contents/table/snack_gfn}

We evaluate whether SNACK can effectively support goal-directed graph generation within the GFlowNet framework, comparing it against the 4-layer graph Transformer following \citet{jain2023multi}. Both SNACK and graph-based baselines use automorphism correction following \citet{kim2025symmetry}. HOMO–LUMO gap optimization assesses whether SNACK can match the performance of a graph policy on a standard, small-graph benchmark (max $N=9$). In addition, we include two similarity-based optimization tasks using albuterol and aripiprazole as reference molecules; these tasks probe the scalability of each method as the trajectory length expand (max $N=16$). 

Table~\ref{tab:final_snkgfn} shows that SNACK is competitive with the graph-based GFlowNet on the HOMO–LUMO gap task and substantially outperforms it on both similarity rewards, while maintaining a unique fraction of $1.0$. In contrast, the graph-based GFlowNet did not reach a unique fraction of 1.0, and its markedly lower Top 100 reward (0.171 versus 0.736 on albuterol) suggests that the graph policy struggles to scale as the action space expands. Notably, results further improve when BFS masking is enforced. This constraint is particularly beneficial for GFlowNet training, as it ensures that every generated sequence possesses a positive backward flow, thereby enhancing overall training stability. 

We also trained GFlowNets directly on SMILES, but they failed to make meaningful progress. This failure is primarily attributed to the inability of SMILES to guarantee structural validity; consequently, early-stage samples were dominated by invalid, zero-reward strings. This shows that SNACK provides a more robust foundation for training GFlowNet models, while SMILES-based models inherently require pretraining to mitigate the complexities of their string representation.

\paragraph*{Conditioned optimization}

\begin{wraptable}{r}{0.4\textwidth} 
\vspace{-1.5em}
\centering
\caption{Conditioned optimization} 
\label{tab:constrained_opt}
\small
\setlength{\tabcolsep}{2pt} 
\begin{tabular}{lccc}
\toprule
Method & Improvement & Similarity & Success \\ 
\midrule
JT-VAE  & $1.91_{\pm 2.0}$ & $0.28_{\pm .1}$ & 97.5\% \\
GCPN    & $4.20_{\pm 1.3}$ & $0.32_{\pm .1}$ & \cellcolor{paperhi!15}\textbf{100\%} \\ 
GraphDF & $5.93_{\pm 0.8}$ & $0.30_{\pm .1}$ &\cellcolor{paperhi!15}\textbf{100\%} \\ 
\midrule
SNACK & \cellcolor{paperhi!15}$\mathbf{6.45}_{\pm 1.6}$ & \cellcolor{paperhi!15}$\mathbf{0.46}_{\pm .15}$ & \cellcolor{paperhi!15}\textbf{100\%} \\ 
\bottomrule
\end{tabular}
\vspace{-1em} 
\end{wraptable}

In this experiment, we focus on optimizing molecular structures around a given fragment, a task of critical importance in drug discovery for lead optimization. Following previous work \citep{jin2018junction,you2018graph}, we constrain the generated molecules to contain one of the 800 molecules with low penalized logP, and train with the conditioned TB objective detailed in Appendix~\ref{app:conditioned gflownet}. Since every prefix of a valid SNACK sequence that ends with a node-type token corresponds to a graph, the task amounts to sequentially appending tokens to the initial string representation of the given fragment. Table~\ref{tab:constrained_opt} demonstrates that SNACK (V+B) effectively optimizes target properties while strictly preserving the required substructure. This suggests that SNACK is fully compatible with principled GFlowNet objectives for conditioned generation, despite its string-based nature.

\paragraph*{Scalability analysis}

\begin{wrapfigure}[14]{r}{0.35\linewidth}
    \vspace{-20pt}
    \centering
    \includegraphics[width=\linewidth]{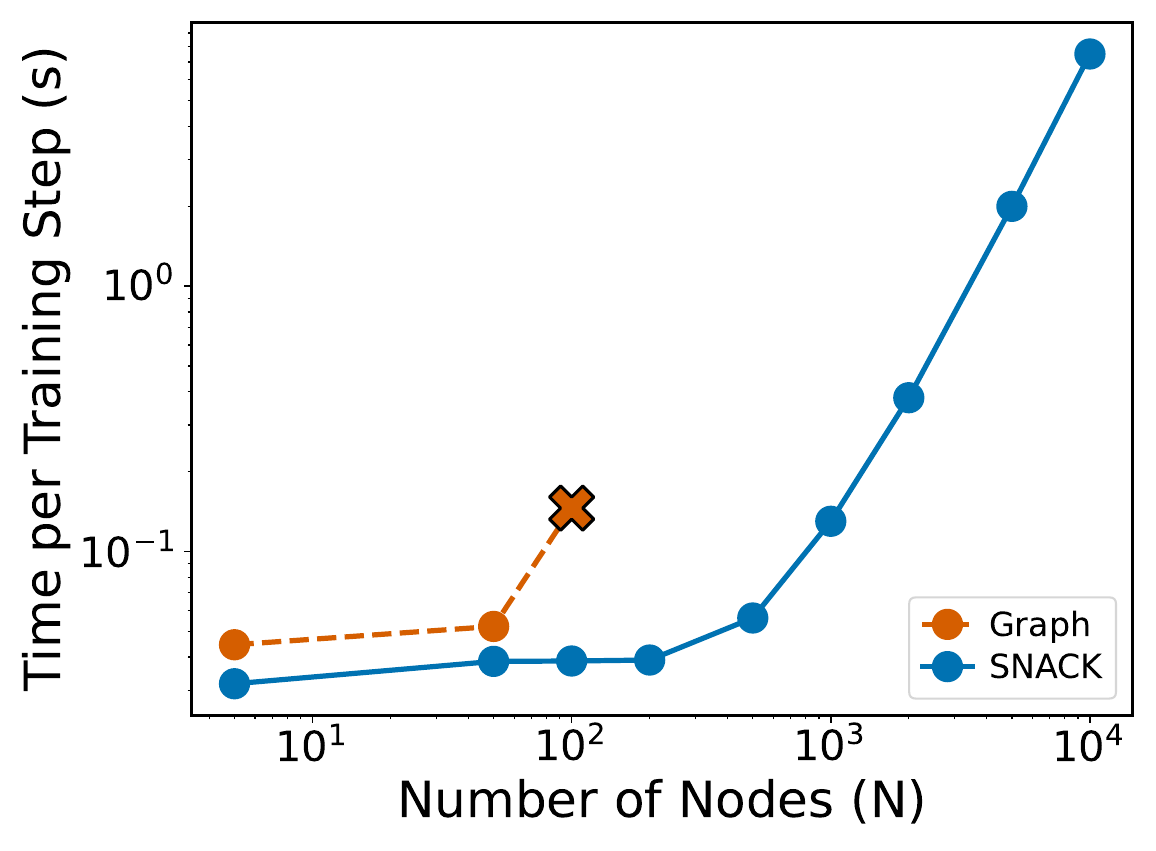}
    \caption{GFlowNet training time comparison. The `×' indicates the maximum scale within GPU memory constraints.}
    \label{fig:training_time}
    \vspace{-6pt}
\end{wrapfigure}

We further evaluate the empirical scalability of SNACK and graph-based GFlowNet by measuring training throughput across molecules of varying sizes, ensuring a fair comparison by maintaining consistent model capacities and hardware configurations. Figure~\ref{fig:training_time} demonstrates that SNACK-GFlowNet exhibits superior computational efficiency as molecular size increases, significantly outperforming the graph-based baseline in terms of training throughput. Additional results regarding inference latency and memory consumption are provided in Appendix~\ref{app:scalability_analysis}. Together, these findings indicate that SNACK enables effective goal-directed GFlowNet learning without the need for graph-native operations, while offering substantial scalability advantages.

\subsection{Ablation studies}

We investigate two alternative design choices for SNACK on molecular datasets. 
Further ablation studies, including investigations into the impact of edge ordering (e.g., left-to-right encoding) and variations in model scale, are provided in Appendix~\ref{app:additional_experimental_results}.

\paragraph*{Different node orderings} 

\begin{wraptable}{r}{0.40\textwidth} 
\vspace{-1.5em}
\centering
\caption{Comparison of FCD scores across different node ordering strategies.}
\label{tab:ordering_comparison}
\small
\begin{tabular}{lcc}
\toprule
 & QM9 & MOSES \\
\midrule
Unif. & $0.203_{\pm 0.013}$ & $6.513_{\pm 0.710}$ \\
CM    & $0.044_{\pm 0.004}$          & $0.113_{\pm 0.010}$ \\
BFS   & \cellcolor{paperhi!15}$\mathbf{0.037}_{\pm 0.003}$ & \cellcolor{paperhi!15}$\mathbf{0.103}_{\pm 0.002}$ \\
\bottomrule
\end{tabular}
\vspace{-1em} 
\end{wraptable}

We evaluate three node ordering strategies: uniform, BFS, and CM. Table~\ref{tab:ordering_comparison} summarizes results. BFS ordering outperforms both uniform and CM orderings on both datasets. We hypothesize that BFS strikes a balance between consistency and diversity of the training data: Uniform ordering produces sequences with substantially different semantics, whereas CM ordering does not provide sufficient diversity. Importantly, this finding contrasts with prior studies \citep{diamant2023improving, jang2023simple} that favor CM ordering. We attribute this discrepancy to two factors. First, the relative advantage of CM ordering is likely dataset-dependent. Second, prior conclusions were often based on evaluation metrics that do not fully capture generative quality. A more comprehensive analysis is provided in Appendix~\ref{app:different node orderings}.

\paragraph*{Different edge orderings}
\begin{wrapfigure}{r}{0.38\linewidth}
    \vspace{-20pt}
    \centering
    \includegraphics[width=\linewidth]{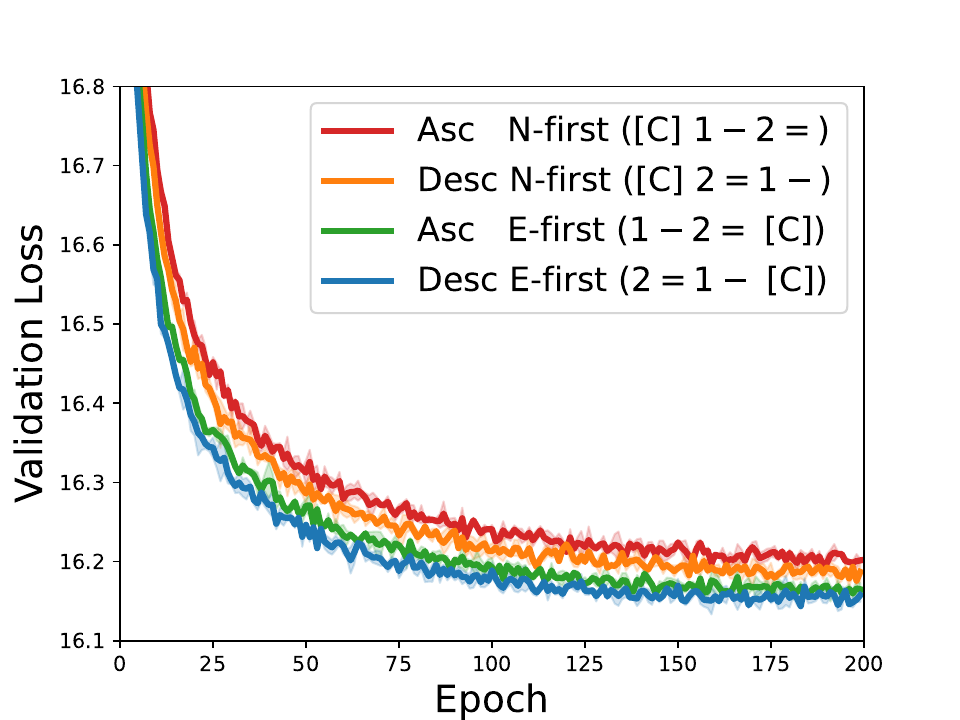}
    \caption{Validation loss of SNACK (None) on QM9 dataset for each block design.}
    \label{fig:qm9_loss}
    \vspace{-10pt}
\end{wrapfigure}
A SNACK string can be decomposed into what we refer to as ``blocks,'' each consisting of a node and its edges to previously generated nodes. For example, \texttt{2=1-[C]} forms a block that connects the current node to two distinct preceding nodes. We evaluate four alternative within-block design choices, including ascending distance ordering (\texttt{1-2=[C]}) and node-first generation (\texttt{[C]2=1-}). Node-first generation is used in GEEL and also common in graph-based methods \citep{li2018learning,you2018graph,shi2020graphaf,kim2025symmetry}. Perhaps surprisingly, however, we find that it diminishes performance consistently in our experiments as shown in Figure~\ref{fig:qm9_loss}. This finding suggests that placing connectivity information before the node token can provide a better inductive bias for autoregressive modeling, justifying the current SNACK representation.

\section{Probing graph structure in learned representations}
\label{sec:internal}

In this section, we use bilinear probing to investigate whether the model’s hidden representations encode underlying graph structure despite the absence of explicit supervision. This is of particular interest for two reasons: First, while prior works often inject structural information directly into models via attention biases \citep{ying2021transformers, ahn2021spanning}, the extent to which a model can internalize such information without these explicit biases merits closer examination. Second, recent studies demonstrate that generative models trained solely on linearized representations can be effective for downstream prediction tasks \citep{chen2025graph, wang2025txgemma}. This raises an important question regarding what structural information is captured by the model during next-token prediction.

Specifically, we assess whether the adjacency matrix and shortest path distance (SPD) matrix can be recovered from the model’s hidden states that correspond to node-type tokens. Let $\mathbf{h}_i^{\ell} \in \mathbb{R}^d$ denote the representation of the $i$-th node at layer $\ell$, with $\mathbf{h}_i^{0}$ being the non-contextual embedding. We normalize and aggregate them via a learned linear combination:
\(
\mathbf{h}_i = \sum_\ell a^{\ell} \mathbf{h}_i^{\ell}/\|\mathbf{h}_i^{\ell}\|,
\)
where $a^{\ell} \in \mathbb{R}$ are learned weights. We then train a bilinear probe parameterized by $\mathbf{W}_K, \mathbf{W}_Q \in \mathbb{R}^{d \times d}$:
\begin{equation}
    \label{eqn:bilinear}
    \mathbf{k}_i = \mathbf{W}_K \mathbf{h}_i \quad
    \mathbf{q}_j = \mathbf{W}_Q \mathbf{h}_j \quad
    \hat{y}_{i,j} = \sigma\!\left(\mathbf{k}_i^\top \mathbf{q}_j\right), \quad
    1 \le i < j \le N
\end{equation}

For edge prediction, $\sigma$ denotes the sigmoid function; for SPD estimation, it is the identity. While alternative normalization schemes exist \citep{peters-etal-2018-deep, tenney2019bert}, we adopt this formulation for its simplicity and effectiveness. Note that we are essentially training a single weight matrix; overparameterizing it with two full-rank matrices was chosen because we observed that it speeds up training \citep{arora2018optimization}.

\begin{figure}[t]
    \centering
    \includegraphics[width=\linewidth]{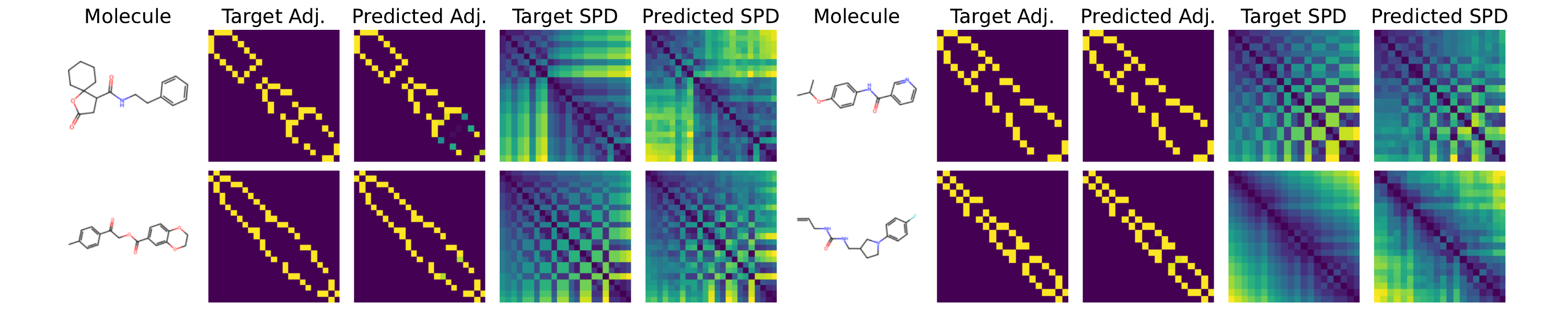}
    \caption{Adjacency and SPD matrices recovered from the model's hidden representations.}    \label{fig:probing}
    \vspace{-5pt} 
\end{figure}

Using Equation~\ref{eqn:bilinear}, we probe whether a SNACK (V+B) model trained on MOSES encodes graph structure in its hidden representations. Figure~\ref{fig:probing} compares model predictions on test molecules with ground truth. The near-perfect reconstruction of adjacency matrices (AUROC: 0.9998) indicates that the SNACK-Transformer encodes local connectivity, while the alignment in SPDs (MAE: 0.9250) suggests that it also captures aspects of the graph's global geometry. This suggests that explicit graph biases are not strictly necessary for capturing complex connectivity; rather, autoregressive pre-training enables the Transformer to internalize the graph’s structure. Table~\ref{tab:probe_results} highlights key performance trade-offs between the two approaches: SNACK excels at edge prediction, whereas SMILES is superior at SPD prediction. This contrast suggests that SNACK prioritizes local connectivity, whereas SMILES representations preserve long-range pairwise distances.

\begin{table}[tbp]
\centering
\caption{Model comparison on probing. ``Random'' denotes a randomly initialized SNACK model.}
\label{tab:probe_results}
\begin{tabular}{l|cc|cc}
\toprule
 & \multicolumn{2}{c|}{Adjacency} & \multicolumn{2}{c}{SPD} \\
Model & Accuracy $\uparrow$ & AUROC $\uparrow$ & MSE $\downarrow$ & MAE $\downarrow$ \\
\midrule
Random & 0.9191 & 0.9362 & 3.0564 & 1.2897 \\
SMILES & 0.9894 & 0.9970 & \cellcolor{paperhi!15}\textbf{1.2722} & \cellcolor{paperhi!15}\textbf{0.8148} \\
SNACK  & \cellcolor{paperhi!15}\textbf{0.9976} & \cellcolor{paperhi!15}\textbf{0.9998} & 1.6659 & 0.9250 \\
\bottomrule
\end{tabular}
\end{table}

\section{Conclusion}
\label{sec:conclusion}
In this work, we introduced SNACK, a simple and efficient string representation for graphs that captures structural information through a construction process analogous to adjacency matrix generation. We showed that SNACK enables the direct enforcement of structural constraints, supports GFlowNet training, and achieves competitive performance across several benchmarks. Furthermore, through probing analyses, we demonstrated that models trained with SNACK can recover meaningful graph structure from their internal representations, even without explicit supervision. A limitation of our approach is that the ordering policy is fixed and manually designed. Important directions for future work include extending SNACK to incorporate stereochemistry and chirality, learning the ordering policy jointly with the model, and adapting the representation for fragment-level generation to better capture higher-level structural patterns.

%% file: contents/table/generic_main.tex
\begin{table*}[hbt]
\centering
\caption{MMD scores for degree (Deg.), clustering (Clus.), and orbit (Orb.) statistics on selected datasets. Results within one standard deviation of the training graphs are highlighted in bold. The baseline results are obtained from \citet{jang2023simple}.}
\label{tab:generic_main}
\setlength{\tabcolsep}{3.5pt}
\small
\begin{tabular}{lccc|ccc|ccc|ccc}
\toprule
 & \multicolumn{3}{c|}{Planar}
 & \multicolumn{3}{c|}{Enzymes}
 & \multicolumn{3}{c|}{SBM}
 & \multicolumn{3}{c}{Grid} \\
\cmidrule(r){2-4} \cmidrule(r){5-7} \cmidrule(r){8-10} \cmidrule(r){11-13}
Method
& Deg. & Clus. & Orb.
& Deg. & Clus. & Orb.
& Deg. & Clus. & Orb.
& Deg. & Clus. & Orb. \\
\midrule

Training 
& 0.001 & 0.004 & 0.000
& 0.010 & 0.019 & 0.009
& 0.015 & 0.001 & 0.037
& 0.000 & 0.000 & 0.000 \\
\midrule

GraphRNN
& 0.005 & 0.278 & 1.254
& 0.017 & 0.062 & 0.046
& \cellcolor{paperhi!15}\textbf{0.006} & 0.058 & 0.079
& 0.011 & \cellcolor{paperhi!15}\textbf{0.000} & 0.021 \\

GDSS
& 0.250 & 0.393 & 0.587
& 0.026 & 0.061 & \cellcolor{paperhi!15}\textbf{0.009}
& 0.496 & 0.456 & 0.717
& 0.111 & 0.005 & 0.070 \\

DiGress
& \cellcolor{paperhi!15}\textbf{0.000} & \cellcolor{paperhi!15}\textbf{0.002} & 0.008
& \cellcolor{paperhi!15}\textbf{0.011} & 0.039 & \cellcolor{paperhi!15}\textbf{0.010}
& \cellcolor{paperhi!15}\textbf{0.006} & 0.051 & 0.058
& 0.016 & \cellcolor{paperhi!15}\textbf{0.000} & 0.004 \\

GEEL
& \cellcolor{paperhi!15}\textbf{0.001} & 0.010 & 0.001
& \cellcolor{paperhi!15}\textbf{0.005} & \cellcolor{paperhi!15}\textbf{0.018} & \cellcolor{paperhi!15}\textbf{0.006}
& 0.025 & 0.003 & \cellcolor{paperhi!15}\textbf{0.026}
& \cellcolor{paperhi!15}\textbf{0.000} & \cellcolor{paperhi!15}\textbf{0.000} & \cellcolor{paperhi!15}\textbf{0.000} \\

\midrule
SNACK
& \cellcolor{paperhi!15}\textbf{0.001} & \cellcolor{paperhi!15}\textbf{0.005} & 0.003
& \cellcolor{paperhi!15}\textbf{0.003} & \cellcolor{paperhi!15}\textbf{0.013} & \cellcolor{paperhi!15}\textbf{0.001}
& \cellcolor{paperhi!15}\textbf{0.018} & 0.013 & \cellcolor{paperhi!15}\textbf{0.039}
& \cellcolor{paperhi!15}\textbf{0.000} & \cellcolor{paperhi!15}\textbf{0.000} & \cellcolor{paperhi!15}\textbf{0.000} \\

\bottomrule
\end{tabular}
\vspace{-12pt}
\end{table*}

%% file: contents/table/mol_table.tex
\begin{table}[bt]
\centering
\caption{Performance on molecular distribution learning. Results within one standard deviation of the best score are highlighted in bold.}
\label{tab:molecule_learning}
\small
\setlength{\tabcolsep}{3pt} 
\renewcommand{\arraystretch}{0.95} 
\begin{tabular}{lccccc|ccccc}
\toprule
& \multicolumn{5}{c|}{QM9} & \multicolumn{5}{c}{MOSES} \\
\cmidrule(lr){2-6} \cmidrule(lr){7-11}
Method 
& Valid $\uparrow$ 
& FCD $\downarrow$ 
& SNN $\uparrow$ 
& Frag $\uparrow$ 
& Scaf $\uparrow$
& Valid $\uparrow$ 
& FCD $\downarrow$ 
& SNN $\uparrow$ 
& Frag $\uparrow$ 
& Scaf $\uparrow$ \\
\midrule

SMILES        
& 0.995 & 0.053 & \cellcolor{paperhi!15}\textbf{0.630} & 0.995 & 0.976
& 0.993 & \cellcolor{paperhi!15}\textbf{0.100} & 0.528 & \cellcolor{paperhi!15}\textbf{1.000} & 0.898 \\

SELFIES       
& \cellcolor{paperhi!15}\textbf{1.000} & \cellcolor{paperhi!15}\textbf{0.037} & 0.625 & 0.995 & 0.975
& \cellcolor{paperhi!15}\textbf{1.000} & 0.119 & 0.514 & \cellcolor{paperhi!15}\textbf{1.000} & 0.896 \\

SAFE          
& 0.969 & 0.113 & 0.611 & 0.992 & 0.956
& 0.989 & 0.258 & 0.496 & 0.999 & 0.877 \\

STGG          
& \cellcolor{paperhi!15}\textbf{1.000} & 0.121 & 0.602 & 0.994 & 0.962
& \cellcolor{paperhi!15}\textbf{1.000} & 0.114 & 0.528 & \cellcolor{paperhi!15}\textbf{1.000} & 0.891 \\

GEEL
& 0.995 & 0.058 & 0.618 & 0.993 & 0.974
& 0.997 & \cellcolor{paperhi!15}\textbf{0.104} & \cellcolor{paperhi!15}\textbf{0.531} & \cellcolor{paperhi!15}\textbf{1.000} & 0.894 \\

G2PT          
& 0.936 & 0.164 & 0.585 & 0.992 & 0.951
& 0.919 & 0.536 & 0.488 & 0.998 & 0.877 \\

\midrule
SNACK (None)
& 0.998 & 0.046 & 0.627 & 0.995 & \cellcolor{paperhi!15}\textbf{0.977}
& 0.996 & 0.106 & 0.516 & \cellcolor{paperhi!15}\textbf{1.000} & 0.893 \\

SNACK (V)     
& \cellcolor{paperhi!15}\textbf{1.000} & 0.040 & 0.628 & 0.994 & \cellcolor{paperhi!15}\textbf{0.977}
& \cellcolor{paperhi!15}\textbf{1.000} & 0.105 & 0.517 & \cellcolor{paperhi!15}\textbf{1.000} & \cellcolor{paperhi!15}\textbf{0.900} \\

SNACK (V+B) 
& \cellcolor{paperhi!15}\textbf{1.000} & \cellcolor{paperhi!15}\textbf{0.037} & 0.627 & \cellcolor{paperhi!15}\textbf{0.996} & \cellcolor{paperhi!15}\textbf{0.977}
& \cellcolor{paperhi!15}\textbf{1.000} & \cellcolor{paperhi!15}\textbf{0.103} & 0.518 & \cellcolor{paperhi!15}\textbf{1.000} & 0.895 \\
\bottomrule
\end{tabular}
\vspace{-8pt}
\end{table}

%% file: contents/table/snack_gfn.tex
\begin{table}[t]
\centering
\caption{Results for HOMO--LUMO gap (Gap), Aripiprazole similarity (Ari), and Albuterol similarity (Alb) optimization tasks. Bold indicates results within one standard deviation of the best.}
\label{tab:final_snkgfn}
\setlength{\tabcolsep}{3pt}
    \renewcommand{\arraystretch}{1.05}
    \footnotesize
    \resizebox{\linewidth}{!}{%
    \begin{tabular}{llcccccc}
    \toprule
    Reward & Method & Avg.\ Reward & Diversity & Top 100 reward & Top 100 div. & Div.\ Top 100 & Uniq.\ Frac. \\
    \midrule
    Gap & Graph
    & $0.984_{\pm 0.010}$
    & $0.958_{\pm 0.003}$
    & \cellcolor{paperhi!15}$\textbf{1.118}_{\pm 0.006}$
    & \cellcolor{paperhi!15}$\textbf{0.948}_{\pm 0.006}$
    & \cellcolor{paperhi!15}$\textbf{1.118}_{\pm 0.006}$
    & $0.982_{\pm 0.016}$ \\
    
    & SNACK (V)
    & $0.988_{\pm 0.001}$
    & $0.957_{\pm 0.001}$
    & \cellcolor{paperhi!15}$\textbf{1.117}_{\pm 0.005}$
    & $0.940_{\pm 0.006}$
    & \cellcolor{paperhi!15}$\textbf{1.117}_{\pm 0.005}$
    & \cellcolor{paperhi!15}$\textbf{1.000}_{\pm 0.000}$ \\
    
    & SNACK (V+B)
    & \cellcolor{paperhi!15}$\textbf{0.997}_{\pm 0.001}$
    & \cellcolor{paperhi!15}$\textbf{0.959}_{\pm 0.000}$
    & \cellcolor{paperhi!15}$\textbf{1.118}_{\pm 0.002}$
    & \cellcolor{paperhi!15}$\textbf{0.947}_{\pm 0.002}$
    & \cellcolor{paperhi!15}$\textbf{1.118}_{\pm 0.002}$
    & \cellcolor{paperhi!15}$\textbf{1.000}_{\pm 0.000}$ \\
    
    \midrule
    Ari & Graph
    & $0.159_{\pm 0.002}$
    & \cellcolor{paperhi!15}$\textbf{0.956}_{\pm 0.002}$
    & $0.204_{\pm 0.006}$
    & \cellcolor{paperhi!15}$\textbf{0.960}_{\pm 0.000}$
    & $0.204_{\pm 0.006}$
    & $0.997_{\pm 0.002}$ \\

    & SNACK (V)
    & $0.187_{\pm 0.000}$
    & $0.942_{\pm 0.000}$
    & $0.259_{\pm 0.003}$
    & $0.939_{\pm 0.001}$
    & $0.259_{\pm 0.003}$
    & \cellcolor{paperhi!15}$\textbf{1.000}_{\pm 0.000}$ \\

    & SNACK (V+B)
    & \cellcolor{paperhi!15}$\textbf{0.292}_{\pm 0.022}$
    & $0.945_{\pm 0.002}$
    & \cellcolor{paperhi!15}$\textbf{0.469}_{\pm 0.009}$
    & $0.916_{\pm 0.001}$
    & \cellcolor{paperhi!15}$\textbf{0.469}_{\pm 0.009}$
    & \cellcolor{paperhi!15}$\textbf{1.000}_{\pm 0.000}$ \\
    
    \midrule
    Alb & Graph
    & $0.116_{\pm 0.004}$
    & \cellcolor{paperhi!15}$\textbf{0.960}_{\pm 0.001}$
    & $0.171_{\pm 0.017}$
    & \cellcolor{paperhi!15}$\textbf{0.960}_{\pm 0.005}$
    & $0.171_{\pm 0.017}$
    & $0.998_{\pm 0.001}$ \\

    & SNACK (V)
    & $0.353_{\pm 0.101}$
    & $0.937_{\pm 0.002}$
    & $0.552_{\pm 0.152}$
    & $0.921_{\pm 0.011}$
    & $0.552_{\pm 0.152}$
    & \cellcolor{paperhi!15}$\textbf{1.000}_{\pm 0.000}$ \\

    & SNACK (V+B)
    & \cellcolor{paperhi!15}$\textbf{0.529}_{\pm 0.007}$
    & $0.930_{\pm 0.000}$
    & \cellcolor{paperhi!15}$\textbf{0.736}_{\pm 0.007}$
    & $0.912_{\pm 0.002}$
    & \cellcolor{paperhi!15}$\textbf{0.736}_{\pm 0.007}$
    & \cellcolor{paperhi!15}$\textbf{1.000}_{\pm 0.000}$ \\
    \bottomrule
    \end{tabular}%
    }
\vspace{-12pt}
\end{table}

%% file: contents/appendix.tex
\section{Additional details on SNACK}
\label{app:additional details on snack}

\subsection{Definition}

In Section~\ref{sec:overview}, we described SNACK through the encoding of an adjacency matrix. Here, we present an independent formal definition of the SNACK representation, starting with its primary structural unit: the \textit{block}.

\begin{definition*}[Block]
    \label{def:block}
    A \textit{block} consists of a single node token preceded by zero or more distance tokens. The distance tokens are integer-valued and must be arranged in strictly descending order. Each distance token is immediately followed by its corresponding edge-type token.
\end{definition*}

To illustrate, consider the SNACK sequence \texttt{[C]1-[C]2-1-[C]}. According to the definition, this sequence is partitioned into three distinct blocks: \texttt{[C]}, \texttt{1-[C]}, and \texttt{2-1-[C]}. This block-based structure ensures that every node carries its own connectivity information to the previously generated subgraph.

\begin{definition*}[SNACK]
    \label{def:snack}
    A \textit{SNACK} string is defined as a concatenation of $N$ blocks, denoted as $[B_1, B_2, \dots, B_N]$. For each \(i\in\{1,\ldots,N\}\), the \(i\)-th block \(B_i\) contains zero or more distinct distance tokens drawn from \(\{1,\ldots,i-1\}\), arranged in strictly descending order.
\end{definition*}

Distance tokens in $i$-th block represent connections to the preceding $i-1$ nodes in the sequence.

\subsection{Valid masking rule}
\label{app:Valid masking rule}

In this subsection, we describe the rules for valid SNACK strings. For simplicity, we focus on SNACK strings that are valid under \textit{connected search ordering}, in which each newly generated node must be connected to at least one previously generated node. This ordering disallows two cases. (1) Disconnected graphs: SNACK represents such graphs by concatenating the strings of each component. For example, \texttt{[C][C]1-[C]} denotes a disconnected graph with two components, \texttt{[C]} and \texttt{[C]1-[C]}. (2) Uniform ordering: under uniform ordering, a three-node graph can be represented as \texttt{[C][C]2-1-[C]}, which corresponds to first generating two disconnected nodes and then connecting them via a third node. In both cases, a node-type token may follow another node-type token. In contrast, in orderings such as BFS, nodes are generated by expanding from previously visited nodes. As a result, a node-type token cannot directly follow another node-type token, allowing only sequences such as \texttt{[C]1-[C]1-[C]} or \texttt{[C]1-[C]2-[C]} for a three-node graph.

Below, we describe the rules for SNACK strings under this connected search ordering. To allow disconnected graphs, node-type tokens are permitted to follow another node-type token directly. BFS ordering can be further enforced using the rules highlighted in \green{green}, which track the \textit{expansion node}, i.e., the node currently being processed.

\begin{itemize}
    \item \textbf{Initialize:} The first token (except for \texttt{[bos]}) is always a node-type token; initialize the number of nodes as $N \leftarrow 0$. \green{Initialize a pointer $P \leftarrow 1$ to track the current expansion node.}
    
    \item \textbf{Node type:} After a node-type token, either a distance token or \texttt{[eos]} follows. Increment $N \leftarrow N + 1$. Distance tokens take values in $\{1, \dots, N\green{-(P - 1)}\}$. If $N$ reaches the maximum number of nodes, only the \texttt{[eos]} token is allowed.
    
    \item \textbf{Distance:} After a distance token, an edge-type token follows. \green{Denote the current distance by $D$. If this is the first distance token in the current block, set $P \leftarrow N - (D - 1)$.}
    
    \item \textbf{Edge type:} An edge-type token can be followed by either a node-type token or a distance token. Distance tokens take values in $\{1, \dots, D - 1\}$ to enforce a descending order. If $D = 1$, no further distance tokens are allowed, and only node-type tokens may follow.
\end{itemize}

\paragraph*{Example} Under BFS ordering, the possible next tokens after the sequence \texttt{[C]1-[C]2-1-[C]} are \{\texttt{1, 2, 3, [eos]}\}. The maximum distance token is \texttt{3} (i.e., the previous distance token \texttt{2} plus 1).

\subsection{The number of distance tokens}

The number of distance tokens (i.e., the maximum bandwidth) can be estimated from the training data. However, the model may generate sequences that use the maximum distance token, for which larger distances could arise when reordering node IDs. To address this, we allocate additional distance tokens beyond the observed maximum. Because the maximum distance is bounded by the total number of nodes, this expansion remains within a strictly finite range.

\subsection{Bijectivity with the adjacency matrix}
\label{app:bijectivity}

In this section, we provide a formal proof of Proposition~\ref{prop:bijectivity}.

\paragraph*{Proposition~\ref{prop:bijectivity}}
    The SNACK representation is bijective with respect to the adjacency matrix.

\begin{proof}
    Let $\mathcal{A}$ be the space of adjacency matrices and $\mathcal{X}$ be the space of valid SNACK sequences. To establish bijectivity, we show that both the encoding function $f: \mathcal{A} \to \mathcal{X}$ and the decoding function $g: \mathcal{X} \to \mathcal{A}$ are injective. 
    
    The injectivity of $f$ is evident from its construction: the deterministic row-by-row traversal of the lower triangle ensures that distinct matrices are mapped to distinct sequences. For the decoding function $g$, consider any $\mathbf{x} \in \mathcal{X}$. $g(\mathbf{x})$ constructs an adjacency matrix $\mathbf{A} \in \mathcal{A}$ as follows:
    
    \begin{enumerate}
        \item $\mathbf{x}$ is uniquely partitioned into $N$ blocks $[B_1, \dots, B_N]$ based on node tokens.
        \item For each block $B_k$, the node token defines the diagonal entry $\mathbf{A}_{k,k}$.
        \item For each distance-edge pair $(d_j, e_j)$ within $B_k$, we set the off-diagonal entries $\mathbf{A}_{k, k-d_j} = \mathbf{A}_{k-d_j, k} = e_j$. 
    \end{enumerate}

    By definition of SNACK, each distance $d_j$ satisfies $1 \leq d_j < k$, which guarantees that the index $k - d_j$ is a valid preceding node index ($1 \leq k - d_j < k$). Since each node token and each distance–edge pair in \(\mathbf x\) corresponds to a unique entry in the lower triangle of \(\mathbf A\), and the descending order of distance tokens ensures a deterministic mapping without overlap, any two distinct sequences $\mathbf{x} \neq \mathbf{x}'$ will result in at least one differing entry in their reconstructed matrices $g(\mathbf{x}) \neq g(\mathbf{x}')$. This establishes the injectivity of $g$.
\end{proof}

As a final note, we emphasize that ambiguous tokens may break the bijectivity of SNACK. For example, if both \texttt{[O]} and \texttt{[OH2]} are processed to represent the same molecule ($\text{H}_2\text{O}$), the one-to-one mapping between sequences and graphs is lost. We evaluate this variant in Appendix~\ref{app:explicit hydrogen}.

\section{Computing random BFS ordering}

Throughout our GFlowNet experiments, we employ a fixed probability distribution for the random BFS ordering, denoted as $q^{\mathrm{ord}}(\pi|G)$. The procedure for computing this probability is detailed in Algorithm~\ref{alg:random_bfs}. Note that the resulting probability is independent of any specific node indexing and depends solely on the graph structure. Thus, node orderings belonging to the same automorphism group yield identical probability outputs.

\begin{algorithm}[h]
\caption{Computation of log-probability of random BFS ordering}
\label{alg:random_bfs}
\begin{algorithmic}[1]
\Require Graph \(G=(V,E)\), BFS ordering \(\pi=(v_1,\dots,v_n)\)
\Ensure \(\log q(\pi|G)\)
\State \(Queue \gets [v_1]\), \quad \(i \gets 2\), \quad \(P \gets n\)
\Comment{\(n\) choices for the BFS root}
\While{\(Queue\) is not empty}
    \State \(u \gets \textsc{PopFront}(Queue)\)
    \State \(C \gets \{w \in Neighbor(u) : w \notin \{v_1,\dots,v_{i-1}\}\} \)
    \Comment{set of newly discovered neighbors}
    \State \(P \gets P \cdot |C|!\)
    \State append \(v_i,\dots,v_{i+|C|-1}\) to \(Queue\)
    \Comment{the next \(|C|\) nodes in the given BFS order}
    \State \(i \gets i + |C|\)
\EndWhile
\State \Return \(-\log P\)
\end{algorithmic}
\end{algorithm}

\section{Computational complexity of trajectory evaluation}
\label{app:computational_complexity}

We compare the computational cost of evaluating forward-policy
probabilities along a complete construction trajectory. This setting
arises when computing trajectory-level training objectives after the
trajectory has been collected. Let $N$ and $M$ denote the
numbers of nodes and edges in the terminal graph, respectively.

\paragraph{Sequence-based evaluation.}
Let $\mathbf{x}_{\leq L}=(x_1,\ldots,x_L)$ be a complete SNACK
representation. Its length is $L=N+2M+O(1)$, where the constant term
accounts for special tokens. A causal Transformer jointly evaluates
$p_\theta(x_\ell | \mathbf{x}_{<\ell})$ for all
$\ell=1,\ldots,L$ in a single forward pass: causal masking ensures
that each prediction depends only on the corresponding preceding
prefix \citep{vaswani2017attention}.
Thus, the intermediate prefixes need not be encoded separately.
Under standard dense self-attention, the resulting arithmetic
complexity is
\begin{equation}
    C_{\mathrm{SNACK}}
    = \Theta(L^2)
    = \Theta\!\left((N+M)^2\right).
    \label{eq:snack_trajectory_complexity}
\end{equation}

\paragraph{Graph-based evaluation.}
Consider a graph construction trajectory $(G_0,\ldots,G_T)$ that
starts from a fixed-size initial graph and adds one node or one edge
at each transition. Its length is $T=\Theta(N+M)$. Write
$n_t=|V(G_t)|$ and $m_t=|E(G_t)|$ for the size of each intermediate
graph. A graph-based policy that recomputes its representations at
each state requires separate encodings of
$G_0,\ldots,G_{T-1}$. For a graph Transformer with dense attention
over all node pairs, evaluating $G_t$ costs $\Theta(n_t^2)$, giving
\begin{equation}
    C_{\mathrm{graph}}
    = \Theta\!\left(\sum_{t=0}^{T-1} n_t^2\right)
    = O(TN^2)
    = O\!\left((N+M)N^2\right).
    \label{eq:graph_trajectory_complexity}
\end{equation}
Once the trajectory is available, the intermediate graphs can be
processed in batches. Such batching permits
parallel execution but does not eliminate their separate encodings
or the associated arithmetic work. Consequently, $T$ counts
state-specific evaluations, not necessarily sequential model calls.

\paragraph{Comparison under edge sparsity.}
Under the assumption $M=O(N)$, both construction lengths are
$\Theta(N)$. Equations~\ref{eq:snack_trajectory_complexity}
and~\ref{eq:graph_trajectory_complexity} then yield quadratic
and cubic costs, respectively. The cubic bound is tight for the
dense-attention graph policy: adding nodes individually from a
fixed-size initial graph entails
$\sum_{t=0}^{T-1} n_t^2=\Omega(N^3)$.
Therefore,
\begin{equation}
    C_{\mathrm{SNACK}}=\Theta(N^2),
    \qquad
    C_{\mathrm{graph}}=\Theta(N^3).
\end{equation}

\section{Applications of graph-to-sequence transition probability}
\label{app:graph-to-seq_transition_probability}

As alluded to in the main text, the ability to compute $q(\mathbf{x}|G)$ offers several utilities, including: (1) training GFlowNets, (2) substructure-conditioned optimization, (3) learning $q$, and (4) computing the marginal log-likelihood. We already discussed GFlowNet training in Section~\ref{sec:training gflownets}. This section elaborates on the remaining applications.

\paragraph*{Substructure-conditioned optimization}
The task is to train a model that optimizes a given molecule by expanding upon its structure. Let $G_0$ be the initial molecular graph. A sequence model optimizes $G_0$ by first converting it into a string representation, $\mathbf{x}_0$, and then continuing the generation process. In many reinforcement learning algorithms, computing the action probability is essential. While the sequence model typically treats $\mathbf{x}_0$ as a fixed starting point, the conversion $G_0 \to \mathbf{x}_0$ is actually governed by the underlying graph-to-sequence dynamics. The methodology for this computation is detailed in Section~\ref{sec:training gflownets}, and we further discuss its application within the GFlowNet framework in Appendix~\ref{app:conditioned gflownet}.

\paragraph*{Variational inference}
While we employ a fixed random BFS for $q(\mathbf{x}|G)$ throughout this work, we observe that the optimal ordering strategy can vary depending on the dataset, as discussed in Appendix~\ref{app:different node orderings}. Therefore, performance could be further enhanced by jointly optimizing $q(\mathbf{x}|G)$ and $p_\theta$ via Equation~\ref{eqn:graph log-likelihood}, following the approach of \citet{chen2021order}.

\paragraph*{Approximating marginal log-likelihood}
The ability to compute the marginal log-likelihood provides several critical advantages for model evaluation and selection. However, exactly computing the log-likelihood is intractable, as it requires marginalizing over all possible node orderings. Instead, we can approximate $p_\theta(G)$ using an importance sampling approach, similar to \citet{kim2025symmetry}:
\begin{equation}
    p_\theta(G) = \sum_{\mathbf{x} \in \mathcal{X}_G} p_\theta(\mathbf{x}) = \mathbb{E}_{\mathbf{x}\sim q(\mathbf{x}|G)}\left [\frac{p_\theta(\mathbf{x})}{q(\mathbf{x}|G)} \right ] \approx \frac{1}{K}\sum_{i=1}^{K}\frac{p_\theta(\mathbf{x}_i)}{q(\mathbf{x}_i|G)}
\end{equation}

where $\mathbf{x}_i$ are sequences sampled from the variational distribution $q(\mathbf{x}|G)$.

\section{Training Objectives for Conditioned GFlowNets}
\label{app:conditioned gflownet}

We present a framework for training GFlowNets in a conditioned setting, where the model generates molecules by expanding upon a predefined scaffold. This approach directly addresses lead optimization, a pivotal task in drug discovery.

Formally, given an initial graph $G_0$, the objective is to train the model such that the conditional probability $p(G|G_0)$ is proportional to the reward function $R(G|G_0)$. Here, $R(G|G_0)$ is defined to assign zero reward to any graph $G$ that is unreachable from $G_0$, effectively constraining the search space to valid expansions of the scaffold.

We decompose the conditional probability $p(G|G_0)$ into a three-stage transition: $G_0 \to \mathbf{x}_0$, $\mathbf{x}_0 \to \mathbf{x}$, and $\mathbf{x} \to G$. The sequence model $p_\theta(\mathbf{x}|\mathbf{x}_0)$ handles the intermediate transitions $\mathbf{x}_0 \to \mathbf{x}$, while the terminating action $\mathbf{x} \to G$ is deterministic. To complete the path, we must account for the initial transition $G_0 \to \mathbf{x}_0$. To this end, we introduce an \textit{initial policy} $p_0(\mathbf{x}_0|G_0)$ that assigns probabilities to each SNACK representation corresponding to $G_0$. Analogous to the variational distribution $q(\mathbf{x}|G)$, the initial policy $p_0(\mathbf{x}_0|G_0)$ is computed as:
\begin{equation}
    p_0(\mathbf{x}_0|G_0) = |\Aut(\mathbf{A}_0)| \cdot p_0^{\mathrm{ord}}(\pi|G_0)
\end{equation}
for any $\pi \in \Aut(\mathbf{A}_0)$, where $\mathbf{A}_0$ is the adjacency matrix encoded by \(\mathbf x_0\), and $p_0^{\mathrm{ord}}$ denotes the ordering policy for the initial step. Incorporating the initial state distribution $\mu(G_0)$, the conditioned Trajectory Balance (TB) objective is formulated as:
\begin{equation}
    \label{eqn:conditioned_tb}
    \big [ 
    \log Z + \log \mu(G_0) + \log p_0(\mathbf{x}_0|G_0) + \log p_\theta(\mathbf{x}|\mathbf{x}_0) - \log q(\mathbf{x}|G) - \log R(G|G_0) \big ]^2.
\end{equation}
In our experiments, the initial states are sampled from a set of 800 molecules under a uniform distribution; thus, $\log \mu(G_0)$ remains constant and can be omitted during optimization. We also set $p_0$ to be the fixed random BFS in the experiment for convenience; however, the initial policy $p_0$ must be learned to ensure that the marginal property $p(G|G_0) = \sum_{\mathbf{x} \in \mathcal{X}_G} p_\theta(\mathbf{x}|\mathbf{x}_0) p_0(\mathbf{x}_0|G_0) \propto R(G|G_0)$ is strictly satisfied. Finally, in our experiment, we employed a VarGrad variant \citep{richter2020vargrad,zhang2023robust} that estimates $\log Z$ for each mini-batch.

\section{Experimental details}
\label{app:experimental details}

\subsection{Model architecture}
For simplicity, we use a unified model architecture across all experiments. Specifically, we train Transformer models \citep{vaswani2017attention} following the LLaMA architecture \citep{touvron2023llama}, with RMSNorm pre-normalization \citep{zhang2019root}, SwiGLU activations \citep{shazeer2020glu}, and rotary positional embeddings \citep{su2024roformer}. For distribution learning tasks, our model has 6 layers, an embedding dimension of 384, and 6 attention heads, resulting in approximately 8.9M parameters. The exact parameter count varies slightly with the tokenizer, as the vocabulary size determines the output projection layer. For goal-directed learning, we use a 4-layer variant.

\subsection{Generic graph generation}

\paragraph*{Dataset}
We evaluate on eight widely used graph generation benchmarks used in \citep{you2018graphrnn, jo2022score, jang2023simple, chen2025graph}: Planar (synthetic planar graphs; 128/32/40), Lobster (tree graphs; 69/20/11), Enzymes (protein tertiary structures from the BRENDA database \citep{schomburg2004brenda}; 410/59/118), SBM (stochastic block model graphs; 128/32/40), Ego (3-hop ego-networks from the Citeseer dataset \citep{sen2008collective}; 529/76/152), and Grid (2D lattice graphs; 70/10/20); community-small (70/10/20); Ego-small (140/20/40), where the numbers denote the train/validation/test splits. We follow the same splits as in \citet{jang2023simple}.

\paragraph*{Baseline}
We report baseline results obtained from \citet{jang2023simple}, including GraphVAE \citep{simonovsky2018graphvae}, GraphRNN \citep{you2018graphrnn}, GRAN \citep{liao2019efficient}, EDP-GNN \citep{niu2020permutation}, GraphGen \citep{goyal2020graphgen}, GraphGen-Redux \citep{bacciu2021graphgen}, GraphAF \citep{shi2020graphaf}, GraphDF \citep{luo2021graphdf}, BiGG \citep{dai2020scalable}, GDSS \citep{jo2022score}, DiGress \citep{vignac2023digress}, GDSM \citep{luo2022fast}, BwR \citep{diamant2023improving}, HGGT \citep{jang2024graph}, and GEEL \citep{jang2023simple}. Training statistics were recomputed to derive the standard deviation. Due to the resulting changes in compatibility thresholds, the specific methods marked in bold differ from the original reports by \citet{jang2023simple}.

\paragraph*{Tokenization}
Given that generic graphs lack edge attributes, we omit edge-type tokens from the SNACK representation. To accommodate graphs with multiple components, we employ a disconnected BFS masking scheme (refer to Appendix~\ref{app:Valid masking rule}).

\paragraph*{Training}
We train all models using random BFS ordering with a batch size of 16 for up to 1M iterations, evaluating every 10k steps. We use a learning rate of $1\times10^{-4}$ with weight decay of 0.01 and apply gradient clipping with a maximum norm of 5.0. Early stopping is employed with a patience of 20 based on validation performance, and in practice training typically converges and stops around 200k iterations. Generic graph experiments are repeated with five different random seeds, using a single GPU (NVIDIA GeForce RTX 3090) and 8 CPU cores.

\subsection{Molecular string generation}

\paragraph*{Tokenization}
For SMILES and SAFE, we adopt character-level tokenization. In the case of SAFE, molecules are first fragmented using the BRICS algorithm \citep{degen2008art}; if fragmentation fails, we fall back to the SMILES representation. For SNACK, STGG, GEEL, and G2PT, we remove aromatic bonds and consider only single, double, and triple bond types. Node types include atom types and charges (e.g., \texttt{[N+]}), and hydrogens are added implicitly after generation.

All tokenizers include special tokens: \texttt{[bos]}, \texttt{[eos]}, \texttt{[pad]}, and \texttt{[unk]}. The unknown token \texttt{[unk]} is introduced because the vocabulary is constructed from the training data using a single representation per molecule. The \texttt{[bos]}, \texttt{[pad]}, and \texttt{[unk]} tokens are used only during training and are masked out at inference time for all tokenizer types (including, e.g., SMILES).

For STGG, GEEL, G2PT and SNACK, we use the following node-type tokens:

\begin{itemize}
    \item QM9: \texttt{[C-]}, \texttt{[C]}, \texttt{[F]}, \texttt{[N+]}, \texttt{[N-]}, \texttt{[N]}, \texttt{[O-]}, \texttt{[O]}
    \item QM9 (with hydrogens): \texttt{[C-]}, \texttt{[CH-]}, \texttt{[CH2]}, \texttt{[CH3]}, \texttt{[CH]}, \texttt{[C]}, \texttt{[F]}, \texttt{[N+]}, \texttt{[N-]}, \texttt{[NH+]}, \texttt{[NH2+]}, \texttt{[NH2]}, \texttt{[NH3+]}, \texttt{[NH3]}, \texttt{[NH]}, \texttt{[N]}, \texttt{[O-]}, \texttt{[OH2]}, \texttt{[OH]}, \texttt{[O]}
    \item MOSES: \texttt{[Br]}, \texttt{[C]}, \texttt{[Cl]}, \texttt{[F]}, \texttt{[N]}, \texttt{[O]}, \texttt{[S]}
    \item MOSES (with hydrogens): \texttt{[Br]}, \texttt{[CH2]}, \texttt{[CH3]}, \texttt{[CH]}, \texttt{[C]}, \texttt{[Cl]}, \texttt{[F]}, \texttt{[NH2]}, \texttt{[NH]}, \texttt{[N]}, \texttt{[OH]}, \texttt{[O]}, \texttt{[S]}
\end{itemize}

In accordance with previous graph generation literature \citep{you2018graph}, molecules were Kekulized to use explicit single, double, and triple bonds instead of aromatic bonds. Generated molecules were subsequently sanitized.

\paragraph*{Baseline implementation}
For GEEL, G2PT, STGG, and SNACK, we use the same set of node-type tokens and add hydrogen atoms implicitly after generation. GEEL and STGG follow their original token masking strategies (grammar masking for GEEL and grammar+valence masking for STGG). For STGG, we use the code provided in their supplementary material, while the other methods are implemented based on the descriptions in their respective papers. For all methods, training sequences are randomly sampled from valid enumerations, as this randomization has been shown to enhance model performance \citep{arus2019randomized}. 


\paragraph*{Training}
We use a batch size of 128, apply no weight decay, optimize with the Adam optimizer \citep{kingma2014adam}, and clip gradients at 5.0 for all datasets. For the QM9 dataset, we train for 200 epochs with a learning rate of 0.0001 and select the best model based on validation loss. For the MOSES dataset, we train for up to 100 epochs with a learning rate of 0.001 and apply early stopping if no improvement in validation loss is observed for 5 consecutive epochs. From the selected model, 30k molecules are sampled for evaluation. Experiments are repeated with three different random seeds. Molecular experiments are conducted using a single GPU (NVIDIA GeForce RTX 3090) and 8 CPU cores. In one of the training runs, the SNACK-Transformer took approximately 10 hours and stopped after 67 epochs on the MOSES dataset. The validation loss is computed on 10k molecules, and evaluation metrics are computed on a test set of 30k molecules.

\paragraph*{Conditioned generation}
We evaluated substructure-conditioned generation using 129 molecules with at most 13 heavy atoms from the standard MOSES test split. For each molecule, we sampled 100 randomized representations with replacement and generated one continuation per representation, resulting in 12,900 attempts per checkpoint. Each complete molecular representation was supplied as a prefix with only the \texttt{[eos]} token removed, and immediate \texttt{[eos]} generation was permitted. We evaluated three pretrained models per method without additional fine-tuning. An extension was successful if the output was valid, connected, contained the conditioning substructure, and had strictly more heavy atoms.


\subsection{Goal-directed learning}
\label{app:goal_directed}

We evaluate SNACK on two complementary molecular generation tasks to assess its performance across different scales and reward landscapes.

\paragraph*{Tokenization} For the HOMO-LUMO gap task, we use node-type tokens obtained from QM9. For two similarity-based tasks, we use the following node-type tokens, derived from ZINC250k: \texttt{[Br]}, \texttt{[C-]}, \texttt{[C]}, \texttt{[Cl]}, \texttt{[F]}, \texttt{[I]}, \texttt{[N+]}, \texttt{[N-]}, \texttt{[N]}, \texttt{[O+]}, \texttt{[O-]}, \texttt{[O]}, \texttt{[P+]}, \texttt{[P]}, \texttt{[S+]}, \texttt{[S-]}, \texttt{[S]}. For the conditioned optimization task, we utilize the following node-type tokens, derived from the fragments used for conditioning: \texttt{[Br]}, \texttt{[C]}, \texttt{[Cl]}, \texttt{[F]}, \texttt{[I]}, \texttt{[N+]}, \texttt{[N-]}, \texttt{[N]}, \texttt{[O+]}, \texttt{[O-]}, \texttt{[O]}, \texttt{[S-]}, \texttt{[S]}.

\paragraph*{HOMO--LUMO Gap Optimization} Following \citet{bengio2021flow} and \citet{jain2023multi}, we use a small molecule generation environment based on the QM9 dataset. Molecules are generated atom-by-atom up to a maximum of 9 heavy atoms. The reward for a completed molecule is the HOMO--LUMO gap predicted by a pre-trained MXMNet model \citep{zhang2020molecular}. 

\paragraph*{Similarity-based Optimization} To evaluate the scalability of SNACK, we conduct similarity-based optimization for larger drug-like molecules. Specifically, we target the generation of molecules similar to \textit{Aripiprazole} (30 heavy atoms) and \textit{Albuterol} (17 heavy atoms). The reward is defined as the Tanimoto similarity between the generated molecule and the target structure, computed using RDKit fingerprints. These tasks are designed to probe the model's capability as the action space and trajectory length increase significantly compared to the QM9 environment.

\paragraph*{Implementation}
For the graph-based GFlowNet baseline, we use the standard open-source implementation.\footnote{\url{https://github.com/recursionpharma/gflownet}} We adopt its default hyperparameters, which are also used in MO-GFN \citep{jain2023multi}: learning rate $1\times10^{-4}$, gradient clipping 10, 4 layers, uniform policy mix 0.001, and reward exponent 32. With this configuration, the baseline collapses on the albuterol task: the unique fraction drops to 0.001 and the generator converges to repeated empty graphs rather than diverse molecules. We therefore performed an additional hyperparameter search on the two similarity tasks and adopted the best non-collapsing setting for the baseline, namely uniform policy mix 0.1, gradient clipping 1.0, and reward exponent 4; the HOMO–LUMO gap task uses the default values. For SNACK, we use a learning rate $1\times10^{-4}$, uniform policy mix 0.1, gradient clipping 1.0, and reward exponent 32 across all tasks. SNACK employs the same transformer architecture as in our distribution learning experiments, with the number of layers reduced from 6 to 4 to align with the graph-based baseline. To ensure a fair comparison, both models use the identical set of atoms.

\paragraph*{Training}
We match the optimizer, batch size, reward setting, and remaining hyperparameters to the standard graph-based GFlowNet as closely as possible. Each training step uses 32 on-policy trajectories and 32 replay-buffer trajectories, and all models are trained with the trajectory balance objective \citep{malkin2022trajectory}. When computing backward probabilities, we apply the automorphism correction of \citet{kim2025symmetry}, so that the trajectory balance objective is unbiased with respect to the underlying graph symmetry. Each model is trained on a single GPU (NVIDIA GeForce RTX 3090) and a single CPU core until training convergence. We then sample 10,000 molecules from the final model and report the mean and standard deviation over three random seeds. 

For SNACK and SMILES, we use the same respective tokenization schemes as in molecular distribution learning. Since multiple SMILES strings can decode to the same molecule, SMILES does not provide a tractable backward distribution over strings given a molecular graph; we therefore approximate the backward term by setting $q(\mathbf{x}|G)=1$ \citep{kim2024genetic}. Despite hyperparameter tuning, the SMILES-based GFlowNet fails to learn from scratch.

\paragraph*{Evaluation metrics}
Evaluation metrics we presented in Table 4 are defined as follows:
\begin{itemize}
    \item \textbf{Avg. Reward}: The average reward over all generated samples.
    \item \textbf{Diversity}: The average pairwise Tanimoto distance among valid generated samples, using Morgan fingerprints (radius 2, 2048 bits).
    \item \textbf{Top 100 reward}: The average reward of the 100 highest-reward samples.
    \item \textbf{Top 100 diversity}: Diversity among the top 100 reward molecules.
    \item \textbf{Diverse Top 100 reward}: The average reward of the top 100 molecules, ensuring that each pair has a Tanimoto similarity below 0.7.
    \item \textbf{Unique fraction}: The fraction of unique molecules among the valid generated samples.
\end{itemize}

\paragraph*{Conditioned optimization}
Following previous studies \citep{jin2018junction, you2018graph}, we define the initial states using a set of 800 ZINC molecules, specifically selected for their low penalized logP scores. We train a 4-layer transformer by optimizing the conditioned TB objective, as explained in Appendix~\ref{app:conditioned gflownet}. The maximum number of nodes for generation is fixed at 40, accounting for the maximum size of initial fragments (37 nodes). The model is trained using the AdamW optimizer with a learning rate of 0.0005 and a batch size of 128. The reward exponent is set to 256. Given that GFlowNet is an off-policy algorithm, we maintain a replay buffer that is updated with 128 new samples every 100 iterations. The model was trained for 30,000 iterations, and we evaluated the checkpoint at 28,000 iterations, which exhibited the highest average reward. For each of the 800 fragments, we generated 10 samples and selected the one with the highest property score improvement. Similarity is measured using Morgan fingerprints (radius 2) with the Tanimoto similarity coefficient. Finally, the success rate is defined as the percentage of molecules that were successfully optimized across the 800 starting fragments. In Table~\ref{tab:constrained_opt}, we compare our model with JT-VAE \citep{jin2018junction}, GCPN \citep{you2018graph}, and GraphDF \citep{luo2021graphdf}, where baseline results are obtained from their respective papers (corresponding to $\delta = 0$).

\subsection{Bilinear probing}

\paragraph*{Training}
We use 160k training examples from MOSES to train the probe parameters $\mathbf{W}_Q$, $\mathbf{W}_K$, and $\{a^{\ell}\}_{\ell=0}^6$. Hidden representations are extracted from a trained 6-layer SNACK (V+B) model and used to train the probe via binary cross-entropy for adjacency prediction and mean squared error for shortest path distance (SPD) prediction. We use the Adam optimizer \citep{kingma2014adam} with a learning rate of 0.001 and gradient clipping with a maximum norm of 1. For each sample, the target matrix is of size $N \times N$, and we train only on its lower triangular part. The batch size is 8. Training is conducted on a single GPU (NVIDIA GeForce RTX 3090). At inference time, the full matrix is reconstructed by symmetrizing the predictions.

\section{Additional experimental results}
\label{app:additional_experimental_results}

\subsection{Full results on generic graphs}

Full results are presented in Tables~\ref{tab:generic_all1} and \ref{tab:generic_all2}, which report MMD scores for degree, clustering, and orbit statistics; for these metrics, lower values indicate superior performance. In the table, N.A. denotes that all generated samples are invalid, and ``--'' denotes unavailable or unreported results. SNACK (BFS) and SNACK (CM) refer to SNACK trained using BFS and CM node orderings, respectively.

Baseline results are obtained from \citet{jang2023simple}. To ensure a consistent comparison, we recomputed the training statistics to derive the standard deviation; results within one standard deviation of the training data are highlighted in bold. Note that the specific methods marked in bold may differ from the original reports due to these updated compatibility thresholds.

\input{contents/table/generic_all1}
\input{contents/table/generic_all2}

\subsection{Additional results on molecular string generation}
\label{app:additional_results_moses}

\subsubsection{Peptide generation}
\label{app:peptides}

We additionally evaluated our model on the Peptides-func dataset, following \citet{diamant2023improving}. Peptides-func presents a more challenging distribution-learning benchmark, as its molecules average 150 atoms (compared to 22 in MOSES). We removed chirality information (corresponding to \texttt{@} symbols in SMILES) because SNACK does not currently support chirality. We used a 10,000 / 2,000 / 3,371 train/validation/test split and trained models for 100 epochs with an early stopping patience of 20, keeping all other hyperparameters identical to those used in the MOSES experiments.

On this benchmark, SNACK demonstrates a clear advantage in validity metrics and outperforms existing baselines on FCD, while remaining competitive on SNN. Compared to SELFIES, the only other method guaranteeing 100\% validity, SNACK achieves superior performance across all metrics. This likely stems from SNACK's compact, unambiguous tokens, which reduce syntactic redundancies and help the model focus on structural patterns.

\begin{table}[thbp]
  \centering
  \caption{Performance on Peptides-func dataset.}
  \label{tab:performance_comparison}
  \vspace{6pt}
  \small
  \begin{tabular}{lccccc}
    \toprule
    Method & Valid $\uparrow$ & FCD $\downarrow$ & SNN $\uparrow$ & Frag $\uparrow$ & Scaf $\uparrow$ \\
    \midrule
    SMILES & $0.820 {\scriptstyle \,\pm\, 0.027}$ & $0.755 {\scriptstyle \,\pm\, 0.071}$ & $\mathbf{0.762 {\scriptstyle \,\pm\, 0.003}}$ & $0.999 {\scriptstyle \,\pm\, 0.000}$ & $0.887 {\scriptstyle \,\pm\, 0.006}$ \\
    SELFIES & $\mathbf{1.000 {\scriptstyle \,\pm\, 0.000}}$ & $0.444 {\scriptstyle \,\pm\, 0.042}$ & $0.715 {\scriptstyle \,\pm\, 0.002}$ & $\mathbf{1.000 {\scriptstyle \,\pm\, 0.000}}$ & $0.887 {\scriptstyle \,\pm\, 0.003}$ \\
    GEEL & $0.781 {\scriptstyle \,\pm\, 0.028}$ & $1.701 {\scriptstyle \,\pm\, 0.331}$ & $0.689 {\scriptstyle \,\pm\, 0.010}$ & $0.998 {\scriptstyle \,\pm\, 0.000}$ & $\mathbf{0.896 {\scriptstyle \,\pm\, 0.004}}$ \\
    \midrule
    SNACK (None) & $0.938 {\scriptstyle \,\pm\, 0.013}$ & $\mathbf{0.422 {\scriptstyle \,\pm\, 0.064}}$ & $0.755 {\scriptstyle \,\pm\, 0.002}$ & $0.999 {\scriptstyle \,\pm\, 0.000}$ & $0.889 {\scriptstyle \,\pm\, 0.001}$ \\
    SNACK (V+B) & $\mathbf{1.000 {\scriptstyle \,\pm\, 0.000}}$ & $\mathbf{0.376 {\scriptstyle \,\pm\, 0.070}}$ & $0.755 {\scriptstyle \,\pm\, 0.005}$ & $\mathbf{1.000 {\scriptstyle \,\pm\, 0.000}}$ & $0.891 {\scriptstyle \,\pm\, 0.003}$ \\
    \bottomrule
  \end{tabular}
\end{table}

\subsubsection{MOSES benchmark}
\input{contents/table/moses}
\input{contents/table/moses2}
The models presented in the main text were kept smaller (6-layer, 384-dimensional, comparable to G2PT$_{\mathrm{small}}$) to facilitate efficient experimentation. To compare our method with previous work, we followed the evaluation protocol described in the MOSES benchmark \citep{polykovskiy2020molecular} more thoroughly, including the dataset split, the number of generated samples (30,000), and the number of independent runs (3). 

We evaluated two model configurations: (1) a 63M-parameter, 6-layer, 1024-dimensional transformer with 8 heads, matching the capacity of STGG \citep{ahn2021spanning}; and (2) a 71M-parameter, 12-layer, 768-dimensional transformer with 12 heads, comparable to G2PT$_{\mathrm{medium}}$ \citep{chen2025graph}. All models were trained using the AdamW optimizer \citep{loshchilov2017decoupled} with a learning rate of $1 \times 10^{-4}$ and a weight decay of 0.01 for a maximum of 100 epochs, with an early-stopping patience of 5 epochs. We refer readers to \citet{polykovskiy2020molecular} for the definitions of each evaluation metric.

The detailed results are provided in Tables~\ref{tab:moses} and \ref{tab:moses2}, with baseline results retrieved from \citet{ahn2021spanning, jang2023simple, chen2025graph}. While some of these comparisons involve diverse architectures making a strict head-to-head evaluation difficult, the SNACK-Transformer combination effectively matches or exceeds the performance reported by existing baselines.

\subsubsection{Explicit hydrogen modeling}
\label{app:explicit hydrogen}
Explicitly including hydrogen atoms in node types (e.g., \texttt{[N]} and \texttt{[NH]}) can violate sequence bijectivity if distinct tokens map to the same molecule after implicit hydrogens are added to satisfy valency. Therefore, we adopted a representation using only single-atom node types. To evaluate the impact of this design choice on distribution learning, we compared our approach with an explicit hydrogen configuration. For SNACK (V+B), the explicit model yielded FCD scores of $0.040 \pm 0.002$ on QM9 and $0.144 \pm 0.031$ on MOSES, whereas our implicit model achieved $0.037 \pm 0.003$ and $0.103 \pm 0.002$, respectively. These results demonstrate that implicit hydrogen modeling does not compromise, and may even enhance, generative performance.

\subsection{Scalability of GFlowNet training}
\label{app:scalability_analysis}
\begin{figure}[t]
    \centering
    \begin{subfigure}[b]{0.32\linewidth}
        \centering
        \includegraphics[width=\linewidth]{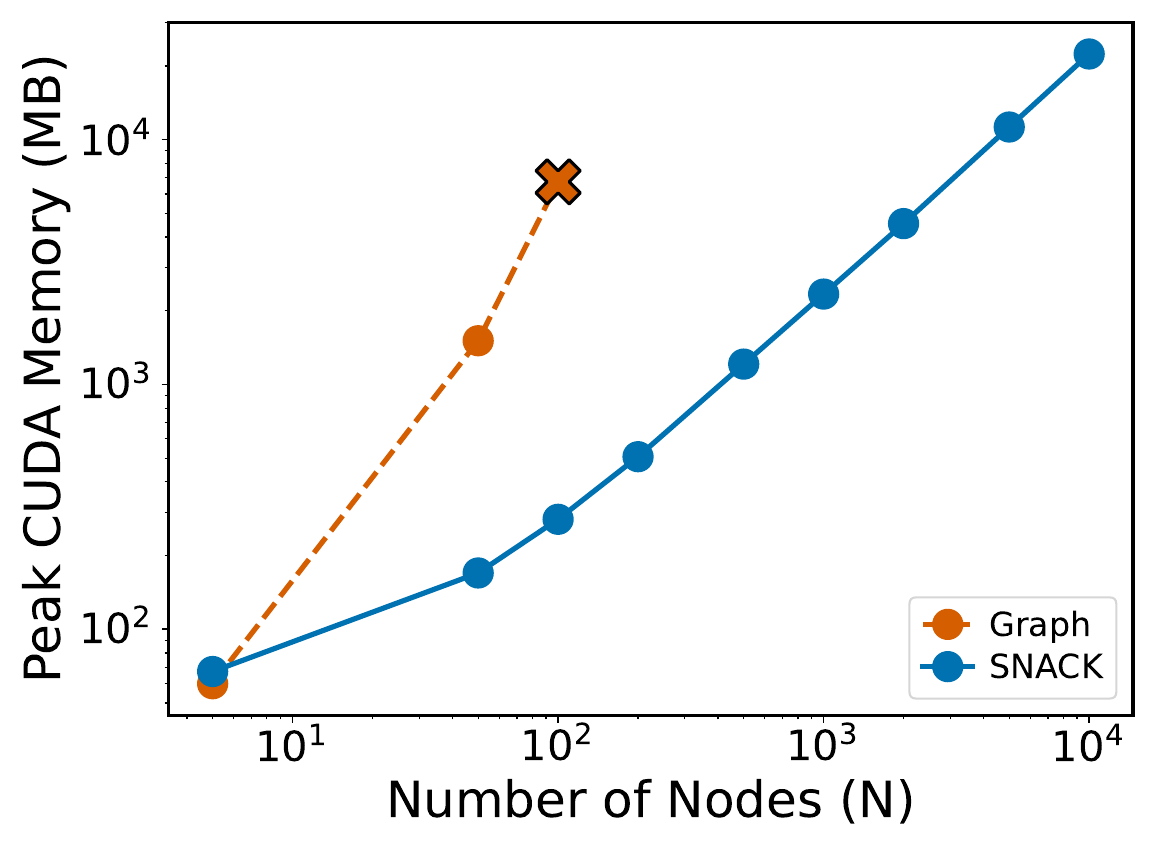}
        \caption{Peak training memory.}
        \label{fig:scalability_train_mem}
    \end{subfigure}
    \hfill
    \begin{subfigure}[b]{0.32\linewidth}
        \centering
        \includegraphics[width=\linewidth]{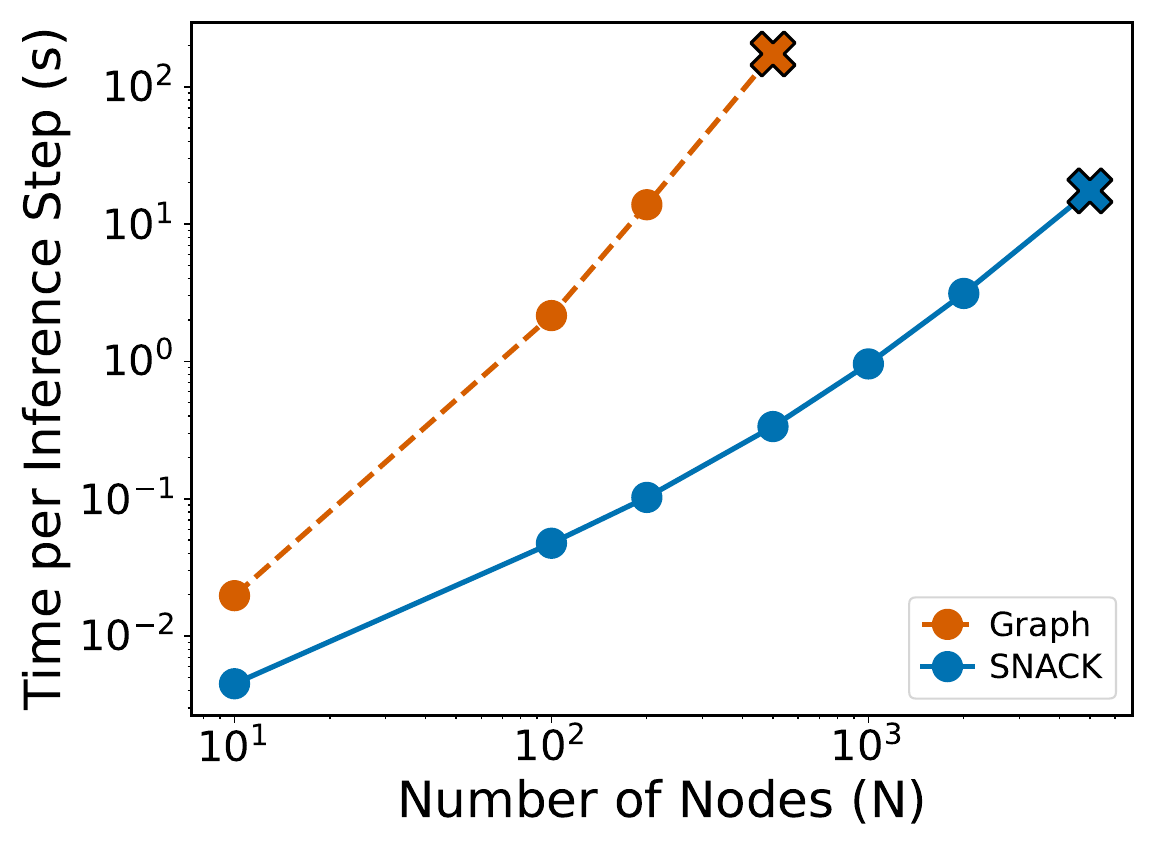}
        \caption{Time per inference step}
        \label{fig:scalability_inf_time}
    \end{subfigure}
    \hfill
    \begin{subfigure}[b]{0.32\linewidth}
        \centering
        \includegraphics[width=\linewidth]{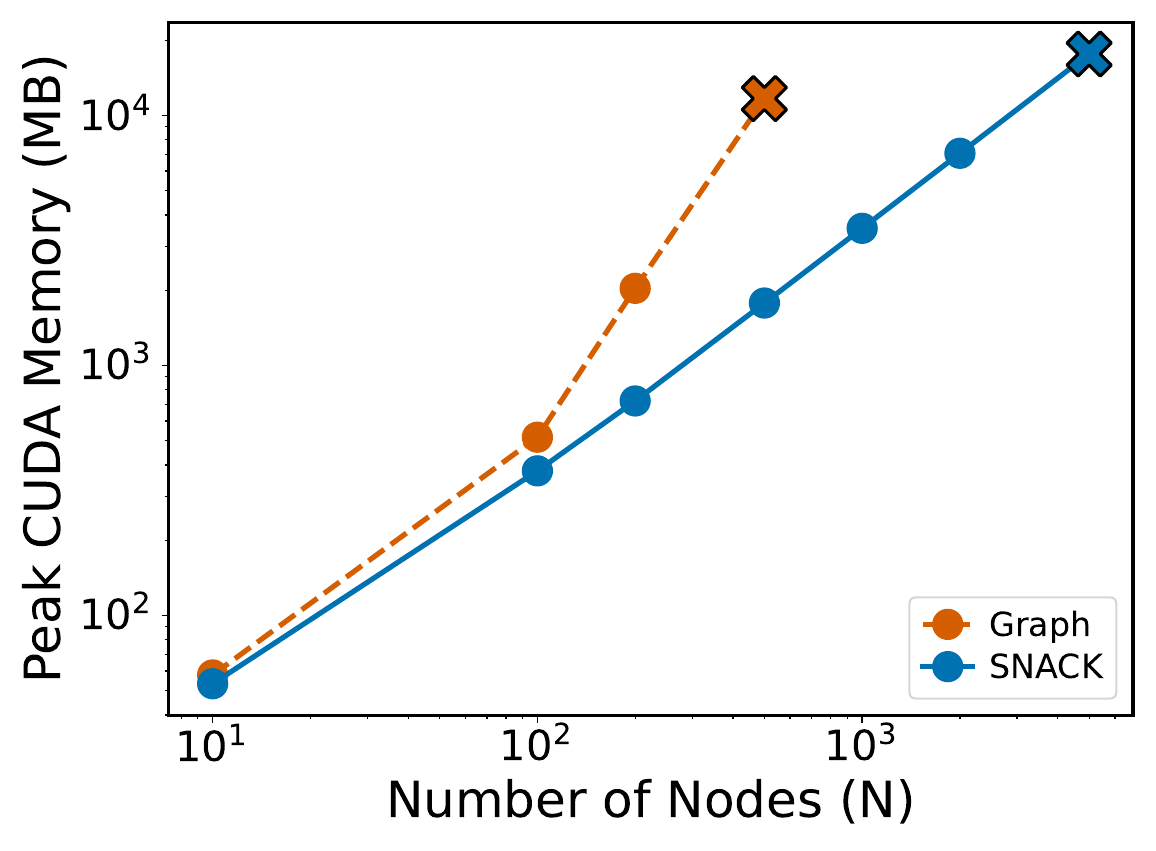}
        \caption{Peak inference memory.}
        \label{fig:scalability_inf_mem}
    \end{subfigure}
    \caption{Scalability of SNACK and the graph GFlowNet versus the number of atoms $N$.}
    \label{fig:scalability_appendix}
\end{figure}

We profile both SNACK and the graph-based GFlowNet on synthetic molecule batches as the number of atoms increases. Training measurements use batch size 8, while inference measurements use batch size 100. For each node size (N) and method, we measure (1) wall-clock time per training step, (2) wall-clock time per inference step, (3) peak CUDA memory during training, and (4) peak CUDA memory during inference. To isolate framework cost, weights are randomly initialized and forward-only generation uses no checkpoints. We use \texttt{torch.cuda.synchronize()} to fence timing measurements and \texttt{torch.cuda.max\_memory\_allocated()} for peak memory, with an extra warm-up pass per size. Following \citep{jain2023multi}, the graph-based GFlowNet uses a 2.2M-parameter graph transformer with 4 layers, embedding dimension 128, and 2 heads. SNACK uses a 2.6M-parameter, 4-layer transformer (embedding dimension 256, 4 heads), trained with full-sequence causal attention and decoded with KV-cached autoregressive generation. All measurements are run on a single GPU (NVIDIA GeForce RTX 3090, 24 GB VRAM) and a single CPU core. When a configuration triggers a CUDA OOM, we record the failure and exclude all larger sizes for that method. The corresponding curves for graph-based GFlowNet and SNACK are therefore truncated at the last size that fit in memory.
The full results are shown in Figure~\ref{fig:scalability_train_mem}, Figure~\ref{fig:scalability_inf_time}, and Figure~\ref{fig:scalability_inf_mem}, together with the main-text training-time result in Figure~\ref{fig:training_time}. Across both training and inference protocols, SNACK is faster and more memory-efficient, with the gap increasing as $N$ grows. This trend is consistent with our complexity analysis.

\subsection{Substructure conditioned generation}
\label{app:substructure_conditioned_generation}

The SNACK framework is exceptionally well-suited for conditional generation, as evidenced by its superior performance. To understand how this works in practice, let us look at a simple example where we generate a complete molecule starting from a partial fragment:
\begin{itemize}
    \item \textbf{SMILES} (\texttt{C(CC)C}): Standard SMILES representations restrict the model to appending new atoms only to the very end of the sequence. This makes it impossible to attach functional groups directly to internal atoms. Additionally, valid prefixes are extremely limited—in this example, only two prefixes (\texttt{C} and \texttt{C(CC)}) are valid—which severely restricts the usable training data for substructure-conditioned generation during pretraining.
    \item \textbf{SNACK} (\texttt{[C]1-[C]1-[C]3-[C]}): In contrast, SNACK gives the model the flexibility to attach new functional groups (such as \texttt{3-[N]}) to \textit{any} previously generated atom within its bandwidth—not just the last one (e.g., directly targeting the second atom). Furthermore, every subsequence that ends with a node-type token acts as a fully valid prefix. In this case, it yields three valid prefixes (\texttt{[C]}, \texttt{[C]1-[C]}, and \texttt{[C]1-[C]1-[C]}), thereby maximizing the utility of the training data during pretraining.
\end{itemize}


\begin{figure}[!th]
    \centering
    \includegraphics[width=\linewidth]{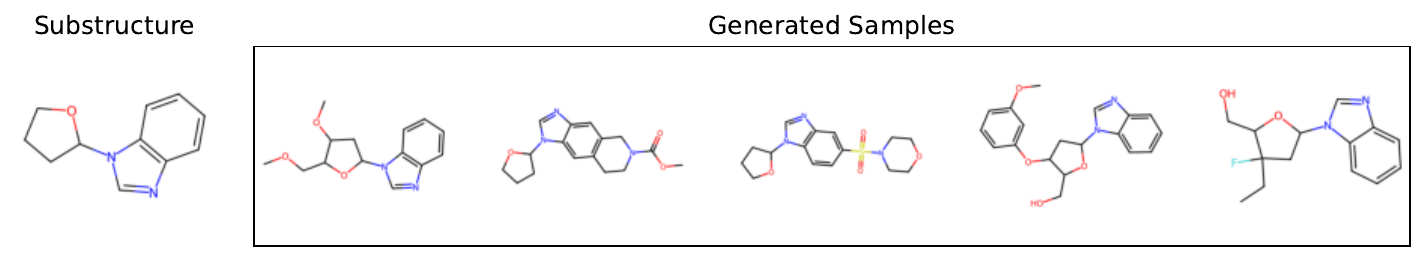}
    \caption{Generated samples conditioned on Maribavir core.}
    \label{fig:superstructure}
    \vspace{-10pt} 
\end{figure}

We show example molecules generated from Maribavir core fragment in Figure~\ref{fig:superstructure}. This fragment is extracted from the drug Maribavir, and used in \citet{noutahi2024gotta} (SMILES: \texttt{c1ccc2c(c1)ncn2[C@@H]1CCCO1}). Utilizing a random BFS-based search, we linearize the given fragments into SNACK representations and proceed with generation from a pretrained SNACK (V+B) checkpoint. This showcases the versatility of the SNACK representation in navigating constrained chemical spaces.


\subsection{Model internals}
\label{app:model_internals}



\begin{table}[th]
\centering
\caption{Reconstruction performance of the bilinear probe.}
\label{tab:probing_results}
\begin{tabular}{llccc}
\toprule
\textbf{Target} & \textbf{Metric} & \textbf{SNACK} & \textbf{SMILES} & \textbf{Random}\\
\midrule
\multirow{2}{*}{Adjacency} & Accuracy $\uparrow$ & 0.9976 & 0.9894 & 0.9191\\
                           & AUROC $\uparrow$    & 0.9998 & 0.9970 & 0.9362\\
\cmidrule(lr){1-5}
\multirow{2}{*}{SPD}       & MSE $\downarrow$    & 1.6659 & 1.2722 & 3.0564 \\
                           & MAE $\downarrow$    & 0.9250 & 0.8148	& 1.2897 \\
\bottomrule
\end{tabular}
\end{table}

\begin{figure}[th]
    \centering
    \includegraphics[width=\linewidth]{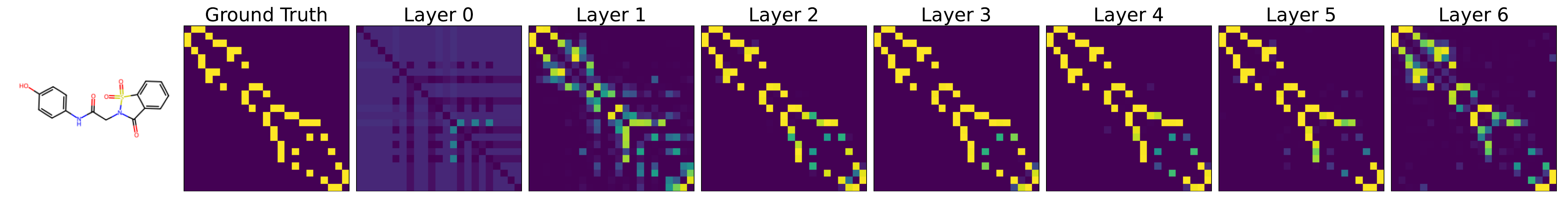}
    \caption{Layer-wise adjacency matrices reconstructed from the hidden representations of the SNACK-Transformer.}
    \label{fig:layer-wise probing}
    \vspace{-5pt} 
\end{figure}

Table~\ref{tab:probing_results} presents the per-molecule reconstruction performance of the bilinear probe on the MOSES dataset. The probe achieves high fidelity in recovering both local connectivity and global geometry, indicating that the learned representations capture essential structural information.

We evaluated the bilinear probe on both the trained SMILES model and an untrained baseline. The trained SMILES model achieves high accuracy in reconstructing the adjacency matrix, indicating that the next-token prediction task generally induces representations that encode graph structural patterns despite the linearized input. However, SNACK achieves higher accuracy on adjacency (edge) prediction, suggesting that SNACK captures more localized information than SMILES.

In contrast, the randomly initialized model reconstructs adjacency matrices that capture only coarse structural trends, yielding blurred probability distributions along the main diagonal—a trivial artifact of BFS ordering, where non-zero entries inherently cluster near the diagonal. Similarly, the recovered SPD matrices merely exhibit averaged patterns rather than fine-grained structural representations.

To further investigate this encoding process, we perform an independent probing of each model layer and visualize the reconstruction of a sample molecule’s adjacency matrix in Figure~\ref{fig:layer-wise probing}. This layer-wise analysis illustrates how structural details are progressively integrated across the network, showing that structural recoverability is most distinct in the intermediate layers.

\section{Analysis of different node ordering strategies}
\label{app:different node orderings}

In this section, we provide a comprehensive examination of how different node ordering policies, denoted as $q(\mathbf{x}|G)$, influence the performance and behavior of graph generative models. While prior literature \citep{diamant2023improving, jang2023simple} has advocated for the use of Cuthill-McKee (CM) ordering, our empirical results on molecular datasets lead to a different conclusion. Before addressing the underlying reasons for this discrepancy, we first outline the theoretical and practical rationale that favors CM ordering.

\subsection{Rationale for CM ordering}

\citet{diamant2023improving} advocated for CM ordering primarily to address two technical challenges: high computational complexity and an expansive output space. In models like GraphRNN \citep{you2018graphrnn}, reduced bandwidth decreases the number of generation steps required per graph, thereby accelerating both training and inference. In SNACK, however, bandwidth dictates the total vocabulary size, which only marginally increases the computational overhead during the transformer's output projection stage. Regarding the output space, SNACK does not inherently hold an advantage over GraphRNN; while GraphRNN extends sequence length to accommodate structural complexity, SNACK expands its vocabulary size. Consequently, CM ordering remains advantageous, as it can reduce the output space from $\mathcal{O}(N^2)$ to $\mathcal{O}(\hat{\varphi} \cdot N)$, where $\hat{\varphi}$ denotes the estimated bandwidth.

\begin{figure}[t]
    \centering
    \includegraphics[width=\linewidth]{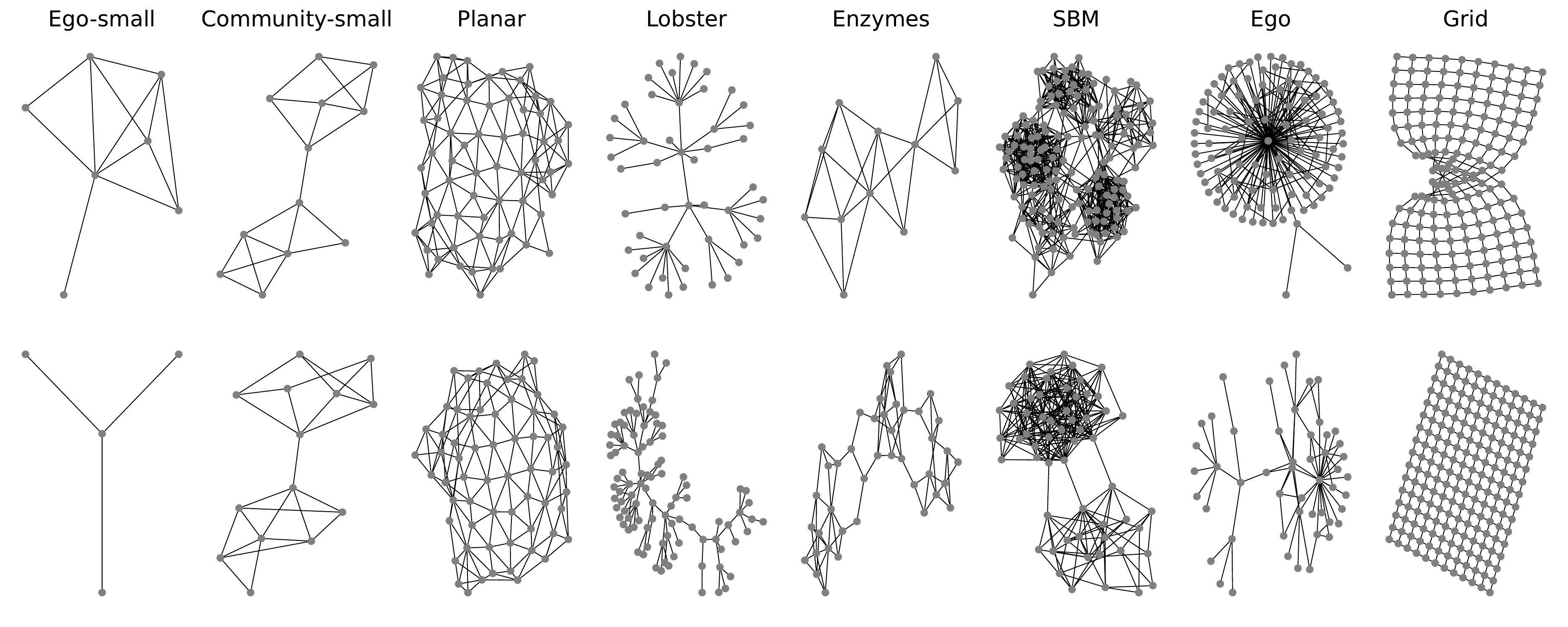}
    \caption{Sample graphs of generic datasets.}
    \label{fig:generic_examples}
    \vspace{-10pt} 
\end{figure}

\begin{figure}[t]
    \centering
    \begin{minipage}{0.52\textwidth}
        \centering
        \makeatletter\def\@captype{table}\makeatother 
        \caption{Maximum bandwidth statistics for generic datasets using BFS versus CM ordering.}
        \label{tab:bandwidth_comparison}
        \small 
        \input{contents/table/CM_BFS}
    \end{minipage}
    \hfill
    \begin{minipage}{0.45\textwidth}
        \centering
        \includegraphics[width=\linewidth]{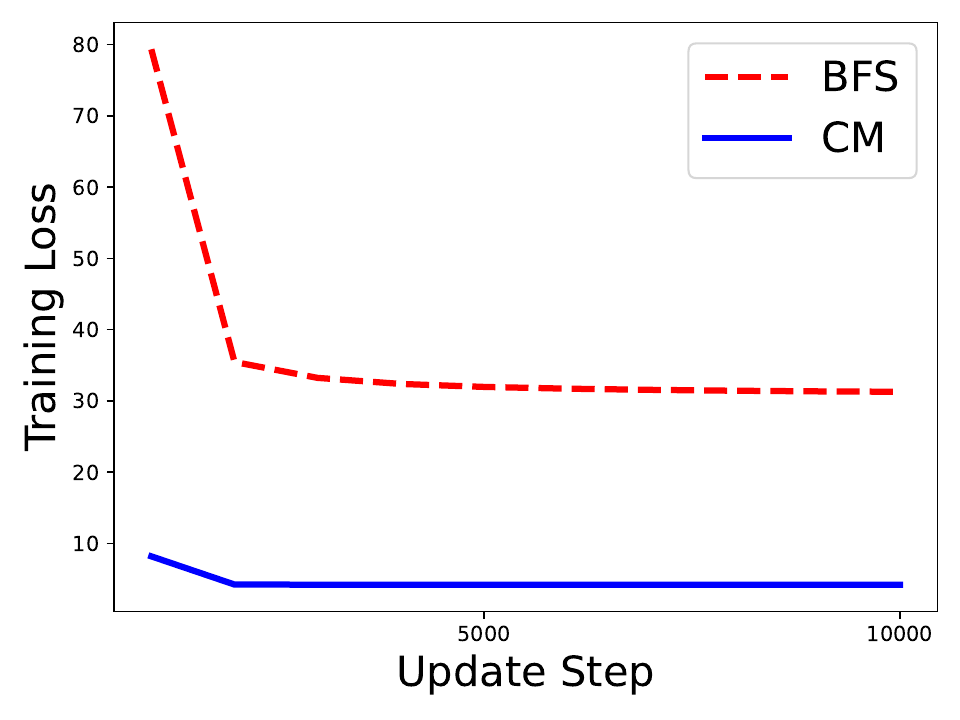}
        \caption{Training loss on Grid dataset.}
        \label{fig:grid_loss}
    \end{minipage}
\end{figure}

\subsection{Optimal ordering policy depends on dataset}

Similar to SNACK, GEEL \citep{jang2023simple} is a string-based method where only the vocabulary size is affected by the graph bandwidth. In their work, experiments on the Grid dataset demonstrated that CM ordering leads to faster convergence. We also observe this trend; furthermore, our results show the superior performance of CM ordering on certain datasets, notably on Community-small and Lobster; see Table~\ref{tab:generic_all1}. These datasets are characterized by their low-degree peripheral structures, where peripheral nodes are clearly distinguished from others. Since CM ordering initiates the traversal from a peripheral node, it yields more consistent sequences for these types of structured graphs. Representative examples of these graphs are visualized in Figure~\ref{fig:generic_examples}. Furthermore, these datasets are characterized by a significant disparity in maximum bandwidth between ordering policies, as evidenced by the results in Table~\ref{tab:bandwidth_comparison}.

In short, our findings suggest that there is no universally superior ordering policy; rather, the optimal ordering is highly dependent on the specific structural characteristics of the dataset. This observation aligns with \citet{chen2021order}, where they demonstrated that jointly training the ordering policy alongside the graph generative model can yield significant performance gains. While CM ordering effectively reduces the output space, our results indicate that the learnability of the resulting sequence—how well the model can capture the structural dependencies within a given budget—is a more critical determinant of generative quality.

\subsection{Analysis of evaluation metrics}

We argue that several evaluation metrics previously employed to compare BFS and CM orderings are inadequate for a rigorous assessment of generative quality.

\paragraph*{Loss and log-likelihood} The negative log-likelihood (NLL) as defined in Equation~\ref{eqn:negative log-likelihood}, or the sequence-level log-likelihood, should not be used as a primary metric for comparison. As discussed in Section~\ref{sec:training objective}, the sequence-level log-likelihood $\log p_\theta(\mathbf{x})$ does not directly equate to the graph-level log-likelihood $\log p_\theta(G)$. Even under the assumption of perfect convergence, these values differ by the entropy of $q(\mathbf{x}|G)$, as illustrated by the following formulation:

\begin{equation}
    \log p_\theta(G) \ge \mathbb{E}_{\mathbf{x} \sim q(\mathbf{x}| G)} \left[ \log p_\theta(\mathbf{x}) \right] + \mathcal{H}(q(\mathbf{x}|G))
\end{equation}

Since different ordering policies (e.g., BFS vs. CM) yield different entropy values $\mathcal{H}(q)$, a direct comparison of their raw loss values is fundamentally misleading. This discrepancy is empirically evidenced by the gap in training loss between BFS and CM orderings shown in Figure~\ref{fig:grid_loss}. Furthermore, we note that the negative log-likelihood (NLL) is frequently computed at the token-level rather than the sequence-level for numerical stability and convenience. In such cases, the inherent differences in loss scales may become less discernible, as the per-token normalization masks the cumulative effect of the ordering's complexity over the entire sequence.
    
\paragraph*{Precision and Recall} While Improved Precision and Recall metrics are widely adopted in the generative modeling literature \citep{kynkaanniemi2019improved} to quantify sample quality and diversity by approximating the data manifold, the metrics used in \citet{diamant2023improving} follow a fundamentally different definition. Specifically, their precision and recall are defined based on the percentage of correct actions (i.e., token-level accuracy) during autoregressive generation. Consequently, these metrics evaluate the model's ability to reconstruct specific training sequences rather than assessing the generated graph distribution. We argue that such action-level metrics are inherently biased toward CM ordering. This bias arises because, for any given intermediate state during generation, there typically exist multiple valid actions that lead to distinct but equally correct trajectories for the same target graph. Since BFS induces a more diverse set of valid trajectories compared to the rigid paths of CM, it may naturally suffer from lower action-level accuracy. 

Consequently, we recommend using metrics that leverage graph-level statistics, such as FCD or MMD, which are invariant to node ordering. Indeed, a closer inspection of the results reported by \citet{diamant2023improving} reveals that BFS ordering performs competitively with CM when evaluated via the MMD metric.

\section{Discussion}
\label{app:discussion}

String-based methods for graph generation have evolved to incorporate the strengths of graph-based approaches. This evolution is primarily driven by the objective of combining the capabilities of Large Language Models (LLMs) with the structural flexibility of graphs. While early molecular graph generation approaches relying on SMILES \citep{weininger1988smiles} often struggled to maintain syntactical validity \citep{segler2018generating}, recent frameworks leverage specialized decoding rules to achieve 100\% validity \citep{krenn2020self,ahn2021spanning}. Expanding beyond the molecular domain, string-based methods have been generalized to generic graphs \citep{jang2023simple, chen2025graph} and further extended to facilitate conditioned generation adhering to specific graph substructures \citep{noutahi2024gotta,chen2025flatten}. As a result, string-based methods have enabled diverse graph generation tasks with superior scalability compared to graph-based approaches \citep{jang2023simple, chen2025flatten}.

However, these advances remain fragmented, as existing string-based methods typically address only a subset of the desired capabilities. Some works do not guarantee strict structural validity \citep{noutahi2024gotta, chen2025graph, chen2025flatten}, others are restricted to molecular graphs and thus lack general applicability \citep{krenn2020self, noutahi2024gotta}, and some do not support flexible conditioned generation in their current form \citep{ahn2021spanning, jang2023simple, chen2025graph}; moreover, evaluations are often confined to distribution learning tasks \citep{ahn2021spanning, jang2023simple, chen2025flatten}, which do not fully capture the challenges of exploration. Although some argue that 100\% validity is of secondary importance since rejection sampling can filter invalid samples \citep{skinnider2024invalid}, in goal-directed settings, such an approach can severely hinder exploration and undermine training stability due to the sparse reward signals from invalid sequences. 

Our paper addresses these gaps by proposing SNACK, a general-purpose string representation that guarantees 100\% structural validity. SNACK provides a flexible framework that supports distribution learning, conditioned generation, and string-based GFlowNet training.

\paragraph*{Limitation}
The SNACK framework currently relies on a fixed node ordering policy and atom-level generation. Although learning the ordering policy and fragment-level generation were not empirically explored in this initial study, SNACK provides the necessary structural foundation to implement such enhancements without architectural modifications, promising further improvements in model expressivity and scalability. Another limitation is that our empirical evaluation relied on the trajectory balance (TB) objective within a specific GFlowNet implementation. While TB provides a robust starting point, its convergence efficiency can be sensitive to reward landscapes and exploration strategies. Future work will investigate alternative training objectives to further stabilize the learning dynamics and enhance the generative diversity of SNACK-based models. Finally, due to constraints in computational resources, we were unable to conduct extensive hyperparameter search. In particular, the high computational cost associated with the graph-based GFlowNet baseline limited the scope of our comparative analysis at larger scales.

\paragraph*{Broader impact} 
By improving graph representation and generative modeling, our work establishes a practical framework that can potentially accelerate molecular design. This approach is particularly valuable for drug discovery and materials science, where efficient exploration of vast chemical spaces is essential. While our model facilitates rapid molecular design, we acknowledge the risk of it being misused to generate harmful or toxic compounds, a challenge that necessitates responsible deployment and screening protocols.

%% file: contents/table/generic_all1.tex
\begin{table}[th]
\centering
\caption{
Results for Ego-small, Community-small, Planar, and Lobster datasets. }
\label{tab:generic_all1}
\setlength{\tabcolsep}{3.2pt}
\small
\begin{tabular}{l|ccc|ccc|ccc|ccc}
\toprule
 & \multicolumn{3}{c|}{Ego-small}
 & \multicolumn{3}{c|}{Community-small}
 & \multicolumn{3}{c|}{Planar}
 & \multicolumn{3}{c}{Lobster} \\
\cmidrule(r){2-4} \cmidrule(r){5-7} \cmidrule(r){8-10} \cmidrule(r){11-13}
 & \multicolumn{3}{c|}{\footnotesize $4 \le |V| \le 18$}
 & \multicolumn{3}{c|}{\footnotesize $12 \le |V| \le 20$}
 & \multicolumn{3}{c|}{\footnotesize $|V| = 64$}
 & \multicolumn{3}{c}{\footnotesize $10 \le |V| \le 100$} \\
\cmidrule(r){2-4} \cmidrule(r){5-7} \cmidrule(r){8-10} \cmidrule(r){11-13}
Method
& Deg. & Clus. & Orb.
& Deg. & Clus. & Orb.
& Deg. & Clus. & Orb.
& Deg. & Clus. & Orb. \\
\midrule

Training
& 0.021 & 0.031 & 0.012
& 0.016 & 0.032 & 0.003
& 0.001 & 0.004 & 0.000
& 0.007 & 0.000 & 0.005 \\

\midrule

GraphVAE
& 0.130 & 0.170 & 0.050
& 0.350 & 0.980 & 0.540
& -- & -- & --
& -- & -- & -- \\

GraphRNN
& 0.090 & 0.220 & \textbf{0.003}
& 0.080 & 0.120 & 0.040
& 0.005 & 0.278 & 1.254
& \textbf{0.000} & \textbf{0.000} & \textbf{0.000} \\

GRAN
& \textbf{0.009} & \textbf{0.038} & \textbf{0.009}
& \textbf{0.005} & 0.142 & 0.090
& \textbf{0.001} & 0.043 & 0.001
& 0.038 & \textbf{0.000} & \textbf{0.001} \\

EDP-GNN
& 0.052 & 0.093 & \textbf{0.007}
& 0.053 & 0.144 & 0.026
& -- & -- & --
& -- & -- & -- \\

GraphGen
& 0.085 & 0.102 & 0.425
& 0.075 & 0.065 & 0.014
& 1.762 & 1.423 & 1.640
& 0.548 & 0.040 & 0.247 \\

GraphGen-Redux
& -- & -- & --
& -- & -- & --
& 1.105 & 1.809 & 0.517
& 1.189 & 1.859 & 0.885 \\

GraphAF
& \textbf{0.030} & 0.110 & \textbf{0.001}
& 0.180 & 0.200 & 0.020
& -- & -- & --
& -- & -- & -- \\

GraphDF
& 0.040 & 0.130 & \textbf{0.010}
& 0.060 & 0.120 & 0.030
& -- & -- & --
& -- & -- & -- \\

BiGG
& \textbf{0.013} & \textbf{0.030} & \textbf{0.005}
& \textbf{0.004} & \textbf{0.005} & \textbf{0.000}
& 0.002 & 0.004 & \textbf{0.000}
& \textbf{0.000} & \textbf{0.000} & \textbf{0.000} \\

GDSS
& \textbf{0.021} & \textbf{0.024} & \textbf{0.007}
& 0.045 & 0.086 & 0.007
& 0.250 & 0.393 & 0.587
& 0.117 & 0.002 & 0.149 \\

DiGress
& \textbf{0.021} & \textbf{0.026} & 0.024
& \textbf{0.012} & \textbf{0.025} & \textbf{0.002}
& \textbf{0.000} & \textbf{0.002} & 0.008
& 0.021 & \textbf{0.000} & \textbf{0.004} \\

GDSM
& -- & -- & --
& \textbf{0.011} & \textbf{0.015} & \textbf{0.001}
& -- & -- & --
& -- & -- & -- \\

BwR
& -- & -- & --
& -- & -- & --
& 0.609 & 0.542 & 0.097
& 0.316 & \textbf{0.000} & 0.247 \\

HGGT
& -- & -- & --
& -- & -- & --
& \textbf{0.000} & \textbf{0.001} & \textbf{0.000}
& \textbf{0.003} & \textbf{0.000} & 0.015 \\

GEEL
& \textbf{0.020} & \textbf{0.035} & \textbf{0.012}
& \textbf{0.020} & \textbf{0.022} & \textbf{0.006}
& \textbf{0.001} & 0.010 & 0.001
& \textbf{0.002} & \textbf{0.000} & \textbf{0.001} \\
\midrule

SNACK (BFS)
& \textbf{0.025} & 0.063 & \textbf{0.010}
& 0.038 & 0.063 & \textbf{0.006}
& \textbf{0.001} & \textbf{0.005} & 0.003
& \textbf{0.009} & \textbf{0.000} & 0.013 \\

SNACK (CM)
& \textbf{0.023} & 0.049 & \textbf{0.013}
& \textbf{0.013} & \textbf{0.035} & \textbf{0.003}
& \textbf{0.001} & 0.031 & 0.012
& \textbf{0.005} & \textbf{0.000} & 0.009 \\

\bottomrule
\end{tabular}
\end{table}

%% file: contents/table/generic_all2.tex
\begin{table}[th]
\centering
\caption{Results for Enzymes, SBM, Ego, and Grid datasets. }
\label{tab:generic_all2}
\setlength{\tabcolsep}{3.2pt}
\small
\begin{tabular}{l|ccc|ccc|ccc|ccc}
\toprule
 & \multicolumn{3}{c|}{Enzymes}
 & \multicolumn{3}{c|}{SBM}
 & \multicolumn{3}{c|}{Ego}
 & \multicolumn{3}{c}{Grid} \\
\cmidrule(r){2-4} \cmidrule(r){5-7} \cmidrule(r){8-10} \cmidrule(r){11-13}
 & \multicolumn{3}{c|}{\footnotesize $10 \le |V| \le 125$}
 & \multicolumn{3}{c|}{\footnotesize $31 \le |V| \le 187$}
 & \multicolumn{3}{c|}{\footnotesize $50 \le |V| \le 399$}
 & \multicolumn{3}{c}{\footnotesize $100 \le |V| \le 400$} \\
\cmidrule(r){2-4} \cmidrule(r){5-7} \cmidrule(r){8-10} \cmidrule(r){11-13}
Method
& Deg. & Clus. & Orb.
& Deg. & Clus. & Orb.
& Deg. & Clus. & Orb.
& Deg. & Clus. & Orb. \\
\midrule

Training
& 0.010 & 0.019 & 0.009
& 0.015 & 0.001 & 0.037
& 0.010 & 0.003 & 0.017
& 0.000 & 0.000 & 0.000 \\
\midrule

GraphVAE
& 1.369 & 0.629 & 0.191
& -- & -- & --
& -- & -- & --
& 1.619 & \textbf{0.000} & 0.919 \\

GraphRNN
& 0.017 & 0.062 & 0.046
& \textbf{0.006} & 0.058 & 0.079
& 0.117 & 0.615 & 0.043
& 0.011 & \textbf{0.000} & 0.021 \\

GRAN
& 0.023 & 0.031 & 0.169
& \textbf{0.011} & 0.055 & 0.054
& 0.026 & 0.342 & 0.254
& 0.001 & 0.004 & 0.002 \\

EDP-GNN
& 0.023 & 0.268 & 0.082
& -- & -- & --
& -- & -- & --
& 0.455 & 0.238 & 0.328 \\

GraphGen
& 0.146 & 0.079 & 0.054
& 1.230 & 1.752 & 0.597
& 0.578 & 1.199 & 0.776
& 1.550 & 0.017 & 0.860 \\

GraphGen-Redux
& 0.456 & 0.035 & 0.251
& -- & -- & --
& 1.088 & 0.702 & 0.155
& -- & -- & -- \\

GraphAF
& 1.669 & 1.283 & 0.266
& -- & -- & --
& -- & -- & --
& -- & -- & -- \\

GraphDF
& 1.503 & 1.283 & 0.266
& -- & -- & --
& -- & -- & --
& -- & -- & -- \\

BiGG
& \textbf{0.010} & \textbf{0.018} & \textbf{0.011}
& 0.029 & 0.003 & \textbf{0.036}
& \textbf{0.010} & 0.017 & \textbf{0.012}
& \textbf{0.000} & \textbf{0.000} & 0.001 \\

GDSS
& 0.026 & 0.061 & \textbf{0.009}
& 0.496 & 0.456 & 0.717
& 0.393 & 0.873 & 0.209
& 0.111 & 0.005 & 0.070 \\

DiGress
& \textbf{0.011} & 0.039 & \textbf{0.010}
& \textbf{0.006} & 0.051 & 0.058
& 0.063 & 0.031 & 0.024
& 0.016 & \textbf{0.000} & 0.004 \\

GDSM
& \textbf{0.013} & 0.088 & \textbf{0.010}
& -- & -- & --
& -- & -- & --
& 0.002 & \textbf{0.000} & \textbf{0.000} \\

BwR
& 0.021 & 0.095 & 0.025
& & N.A. & 
& & N.A. & 
& 0.385 & 1.187 & 0.083 \\

HGGT
& \textbf{0.005} & \textbf{0.017} & \textbf{0.000}
& \textbf{0.017} & 0.014 & 0.076
&  & N.A. & 
& \textbf{0.000} & \textbf{0.000} & \textbf{0.000} \\

GEEL
& \textbf{0.005} & \textbf{0.018} & \textbf{0.006}
& 0.025 & 0.003 & \textbf{0.026}
& 0.053 & 0.017 & \textbf{0.016}
& \textbf{0.000} & \textbf{0.000} & \textbf{0.000} \\

\midrule
SNACK (BFS)
& \textbf{0.003} & \textbf{0.013} & \textbf{0.001}
& \textbf{0.018} & 0.013 & \textbf{0.039}
& 0.071 & 0.122 & 0.034
& \textbf{0.000} & \textbf{0.000} & \textbf{0.000} \\

SNACK (CM)
& \textbf{0.006} & 0.023 & \textbf{0.005}
& 0.134 & 0.168 & 0.069
& 0.050 & 0.256 & \textbf{0.019}
& \textbf{0.000} & \textbf{0.000} & \textbf{0.000} \\

\bottomrule
\end{tabular}
\end{table}

%% file: contents/table/moses.tex
\begin{table*}[th]
\centering
\caption{First set of experimental results for the MOSES benchmark.}
\label{tab:moses}
\vspace{0.1cm}
\resizebox{\textwidth}{!}{
\begin{tabular}{l ccc ccc cc}
\toprule
\textsc{Method} & \textsc{Valid} & \textsc{Unique1} & \textsc{Unique2} & \textsc{IntDiv} & \textsc{IntDiv2} & \textsc{Filters} & \textsc{Novel} \\
\midrule
\textsc{Training Dataset} & 1.0000 & 1.0000 & 1.0000 & 0.8567 & 0.8508 & 1.0000 & 0.0000 \\
\midrule
HMM \citep{polykovskiy2020molecular} & 0.0760 & 0.6230 & 0.5671 & 0.8466 & 0.8104 & 0.9024 & \textbf{0.9994} \\
NGRAM \citep{polykovskiy2020molecular} & 0.2376 & 0.9740 & 0.9217 & \textbf{0.8738} & \textbf{0.8644} & 0.9582 & 0.9694 \\
\textsc{Combinatorial} \citep{liu2017break} & \textbf{1.0000} & 0.9983 & 0.9909 & 0.8732 & 0.8666 & 0.9557 & 0.9878 \\
\textsc{CharRNN} \citep{segler2018generating} & 0.9748 & \textbf{1.0000} & 0.9994 & 0.8562 & 0.8503 & 0.9943 & 0.8419 \\
AAE \citep{polykovskiy2020molecular} & 0.9368 & \textbf{1.0000} & 0.9994 & 0.8557 & 0.8499 & 0.9960 & 0.7931 \\
VAE \citep{polykovskiy2020molecular} & 0.9767 & \textbf{1.0000} & 0.9984 & 0.8558 & 0.8498 & 0.9970 & 0.6949 \\
JT-VAE \citep{jin2018junction} & \textbf{1.0000} & \textbf{1.0000} & \textbf{0.9996} & 0.8551 & 0.8493 & 0.9760 & 0.9143 \\
\textsc{LatentGAN} \citep{prykhodko2019novo} & 0.8966 & \textbf{1.0000} & 0.9968 & 0.8565 & 0.8505 & 0.9735 & 0.9498 \\
STGG \citep{ahn2021spanning} & \textbf{1.0000} & \textbf{1.0000} & 0.9987 & 0.8556 & 0.8496 & \textbf{0.9976} & 0.6727 \\
GEEL \citep{jang2023simple} & 0.9999 & - & - & - & - & - & - \\
G2PT$_{\mathrm{large}}$ \citep{chen2025graph} & 0.9720 & \textbf{1.0000} & - & - & - & 0.9890 & 0.7940 \\
\midrule
SNACK (6-layer) & \textbf{1.0000} & 0.9997 & \textbf{0.9996} & 0.8568 & 0.8508 & 0.9964 & 0.7660 \\
SNACK (12-layer) & \textbf{1.0000} & \textbf{1.0000} & 0.9993 & 0.8571 & 0.8511 & 0.9975 & 0.7002 \\
\bottomrule
\end{tabular}
}
\end{table*}

%% file: contents/table/moses2.tex
\begin{table*}[th]
\centering
\caption{Second set of experimental results for the MOSES benchmark.}
\label{tab:moses2}
\vspace{0.1cm}
\resizebox{\textwidth}{!}{
\begin{tabular}{l cc cc cc cc}
\toprule
& \multicolumn{2}{c}{\textsc{FCD} ($\downarrow$)} & \multicolumn{2}{c}{\textsc{SNN} ($\uparrow$)} & \multicolumn{2}{c}{\textsc{Frag} ($\uparrow$)} & \multicolumn{2}{c}{\textsc{Scaf} ($\uparrow$)} \\
\cmidrule(lr){2-3} \cmidrule(lr){4-5} \cmidrule(lr){6-7} \cmidrule(lr){8-9}
\textsc{Method} & \textsc{Test} & \textsc{TestSF} & \textsc{Test} & \textsc{TestSF} & \textsc{Test} & \textsc{TestSF} & \textsc{Test} & \textsc{TestSF} \\
\midrule
\textsc{Training Dataset} & 0.0080 & 0.4755 & 0.6419 & 0.5859 & 1.0000 & 0.9986 & 0.9907 & 0.0000 \\
\midrule
HMM \citep{polykovskiy2020molecular} & 24.466 & 25.431 & 0.3876 & 0.3795 & 0.5754 & 0.5681 & 0.2065 & 0.0490 \\
NGRAM \citep{polykovskiy2020molecular} & 5.5069 & 6.2306 & 0.5209 & 0.4997 & 0.9846 & 0.9815 & 0.5302 & 0.0977 \\
\textsc{Combinatorial} \citep{liu2017break} & 4.2375 & 4.5113 & 0.4514 & 0.4388 & 0.9912 & 0.9904 & 0.4445 & 0.0865 \\
\textsc{CharRNN} \citep{segler2018generating} & 0.0732 & 0.5204 & 0.6015 & 0.5649 & 0.9998 & 0.9983 & 0.9242 & \textbf{0.1101} \\
AAE \citep{polykovskiy2020molecular} & 0.5555 & 1.0572 & 0.6081 & 0.5677 & 0.9910 & 0.9905 & 0.9022 & 0.0789 \\
VAE \citep{polykovskiy2020molecular} & 0.0990 & 0.5670 & 0.6257 & 0.5783 & 0.9994 & 0.9984 & 0.9386 & 0.0588 \\
JT-VAE \citep{jin2018junction} & 0.3954 & 0.9382 & 0.5477 & 0.5194 & 0.9965 & 0.9947 & 0.8964 & 0.1009 \\
\textsc{LatentGAN} \citep{prykhodko2019novo} & 0.2968 & 0.8281 & 0.5371 & 0.5132 & 0.9986 & 0.9972 & 0.8867 & 0.1072 \\
STGG \citep{ahn2021spanning} & 0.0680 & \textbf{0.5032} & \textbf{0.6359} & \textbf{0.5851} & 0.9998 & 0.9984 & 0.9416 & 0.0389 \\
GEEL \citep{jang2023simple}  & 0.1603 & - & 0.6310 & - & 0.9996 & - & 0.8622 & - \\
G2PT$_{\mathrm{large}}$ \citep{chen2025graph}  & 1.0200 & - & 0.5500 & - & - & - & 0.0290 & - \\
\midrule
SNACK (6-layer) & 0.0619 & 0.5050 & 0.6169 & 0.5731 & \textbf{1.0000} & \textbf{1.0000} & 0.9388 & 0.0798 \\
SNACK (12-layer) & \textbf{0.0611} & 0.5257 & 0.6274 & 0.5775 & \textbf{1.0000} & \textbf{1.0000} & \textbf{0.9432} & 0.0590 \\
\bottomrule
\end{tabular}
}
\end{table*}

%% file: contents/table/CM_BFS.tex

\begin{tabular}{lcc}
\toprule
Dataset         & BFS & CM  \\ \midrule
Community-small & 15  & 8       \\
Ego-small       & 16  & 15      \\
Planar          & 33  & 27      \\
Lobster         & 60  & 49      \\
Enzymes         & 26  & 24      \\
SBM             & 120 & 121    \\
Ego             & 284 & 261    \\ 
Grid            & 39  & 19     \\ \bottomrule
\end{tabular}